\documentclass{article}
\usepackage{preprint,times}
\usepackage[T1]{fontenc}

\usepackage{graphicx}
\usepackage{booktabs}
\usepackage{amsmath,amssymb}
\usepackage{array}
\usepackage{tabularx}
\usepackage{hyperref}
\usepackage{url}
\usepackage{enumitem}
\usepackage{etoc}
\usepackage[english]{babel}
\usepackage[protrusion=true,expansion=true,kerning=true,babel=true,final]{microtype}

\usepackage{amsmath,amsfonts,bm}

\def\eqref#1{equation~\ref{#1}}

\def\1{\bm{1}}

\DeclareMathAlphabet{\mathsfit}{\encodingdefault}{\sfdefault}{m}{sl}
\SetMathAlphabet{\mathsfit}{bold}{\encodingdefault}{\sfdefault}{bx}{n}

\title{PHASE: A Physiology-Guided Hierarchical\\
Foundation Model for Intracranial EEG}
\author{%
Yipeng Zhang$^{1}$ \quad
Chenda Duan$^{1}$ \quad
Yuanyi Ding$^{1}$ \quad
Tianyi Wang$^{1}$ \quad
Atsuro Daida$^{2}$ \\
\textbf{Masaki Izumi}$^{2}$ \quad
\textbf{Yuta Tanoue}$^{2}$ \quad
\textbf{Naoto Kuroda}$^{3}$ \quad
\textbf{Shaun A.~Hussain}$^{2}$ \quad
\textbf{Nishant Sinha}$^{4}$ \\
\textbf{Eishi Asano}$^{3}$ \quad
\textbf{Hiroki Nariai}$^{2}$ \quad
\textbf{Vwani Roychowdhury}$^{1}$\thanks{Corresponding author: \href{mailto:vwani@ucla.edu}{\texttt{\textless vwani@ucla.edu\textgreater}}.} \\
\\
\parbox[t]{0.9\textwidth}{
$^{1}$UCLA Samueli School of Engineering\\
$^{2}$UCLA Mattel Children’s Hospital, David Geffen School of Medicine\\
$^{3}$Children's Hospital of Michigan, Wayne State University School of Medicine\\
$^{4}$University of Pennsylvania\\
}
}

\begin{document}

\maketitle
\etocdepthtag.toc{mtchapter}

\begin{abstract}
Clinicians and neuroscientists have long analyzed intracranial electroencephalography (iEEG) through directly measurable physiological characteristics, which carry much of the information that downstream tasks depend on. Recent iEEG foundation models learn by reconstructing or predicting their inputs, which leaves the retention of these characteristics implicit. They are also evaluated mainly on cognitive decoding and a narrow clinical task, i.e., seizure detection. On a broad, clinically relevant benchmark such as Omni-iEEG, they remain below task-specific models when used frozen. We introduce PHASE, a physiology-guided foundation model that makes these characteristics explicit learning targets, pairing them with masked latent prediction in a temporal stage (PHASE-T) within each channel and a spatiotemporal stage (PHASE-ST) across synchronized channels. PHASE is pretrained on heterogeneous recordings from 222 participants at nine clinical sites. On all five Omni-iEEG clinical tasks, frozen PHASE-T outperforms every evaluated foundation model by up to 31\%, and fine-tuned PHASE-T surpasses the task-specific models, setting a new state of the art. PHASE-T benefits from physiological supervision, outperforming variants trained with latent prediction alone or auxiliary waveform reconstruction on every task in matched ablations. PHASE-T generalizes to unseen institutions, outperforming the compared models with few or no local labels. PHASE-ST further improves seizure-onset-zone identification over PHASE-T and, when frozen, decodes sound volume and pitch on BrainTreebank better than published models. Beyond task performance, PHASE learns to encapsulate the physiological characteristics clinicians recognize, from seizure onset and its propagation to anatomical region identity, even though its pretraining contains no ictal recordings or anatomical labels.
\end{abstract}

\section{Introduction}
\label{sec:introduction}

Intracranial electroencephalography (iEEG) provides a direct view of brain activity for both clinical assessment and cognitive neuroscience.
In clinical practice, it helps identify seizure-generating tissue and localize cortical regions supporting essential functions such as language and movement~\citep{rosenow2001presurgical,jayakar2016diagnostic,kambara2018presurgical}.
In cognitive research, recordings during naturalistic listening reveal neural responses to speech and language~\citep{wang2024brain,zada2025podcast}.
Supporting these applications calls for representations that preserve information about both pathological activity and normal brain function across diverse recording settings.

Measurable waveform and spectral properties have long guided the clinical interpretation and cognitive analysis of iEEG.
Clinical criteria identify interictal discharges by waveform shape and describe seizure patterns through changes in frequency, amplitude, and morphology~\citep{kane2017glossary}.
These properties carry information across temporal scales: brief physiological and pathological high-frequency oscillations (HFOs) exhibit characteristic waveform and spectral features~\citep{zhang2022physiologicalhfo,zhang2025pathologicalhfo,daida2026developmental}, whereas ongoing rhythms vary with brain state and cortical region~\citep{andrillon2011sleep,frauscher2018atlas}.
Regional deviations from normative band power can help localize epileptogenic tissue~\citep{taylor2022normative}.
In auditory cortex, high-frequency power indexes local population activity~\citep{lachaux2012high}, and its amplitude tracks sentence onsets, intonation, and sound intensity~\citep{hamilton2018spatial,tang2017intonational,potes2012dynamics}.
This motivates using directly measurable physiological properties to learn representations that preserve the information needed for diverse downstream iEEG tasks.

Recent iEEG foundation models learn temporal and spectral structure from unlabeled recordings through reconstruction and prediction~\citep{wang2023brainbert,zhang2023brant,carzaniga2026mvpformer}.
These objectives leave the retention of this physiological information implicit.
This information is expressed at several scales: brief events are interpreted within ongoing activity, and each channel in relation to the others~\citep{frauscher2024ieeg}.
Electrode placement and coverage are also participant-specific~\citep{jayakar2016diagnostic}, and acquisition practices and recording contexts differ across institutions, so representations are most broadly useful when they depend on neither a fixed channel layout nor electrode coordinates.
The challenge is therefore to learn representations that capture these temporal and spatial relationships in heterogeneous recordings without relying on scarce expert labels.

To address these challenges, we introduce \textbf{PHASE} (\textbf{P}hysiology-Guided \textbf{H}ierarchical \textbf{A}rchitecture for \textbf{S}ignal Understanding in \textbf{E}lectrophysiology), a two-stage foundation model that makes measurable electrophysiology an explicit learning target across temporal and spatial scales.
\textbf{PHASE-T} (PHASE-\textbf{T}emporal) pairs masked latent prediction, which predicts an exponential-moving-average (EMA) teacher's embeddings of the unmasked signal at masked positions~\citep{assran2023jepa}, with physiological targets measured directly from the signal.
Its fine--coarse--fine hierarchy connects local waveform structure with context from inputs of up to one minute.
\textbf{PHASE-ST} (PHASE-\textbf{S}patio\textbf{T}emporal) extends these frozen temporal representations across synchronized channels, adding targets for measured cross-channel relationships, and requires neither electrode coordinates nor fixed channel identities.
Our contributions can be summarized as follows:
\begin{enumerate}[leftmargin=*]
\item \textbf{Physiological supervision leads to more informative representations across temporal and spatial scales.} PHASE turns measurable physiology into learning targets, within each channel and across synchronized channels, while connecting waveform structure to minute-scale context. In matched ablations, PHASE-T's physiological supervision outperforms waveform reconstruction on every task and raises HFO and sleep macro-F1 by 30\% and 37\% over latent prediction (Table~\ref{tab:mechanism-ablations}).
\item \textbf{Diverse pretraining and superior generalization to unseen institutions.} PHASE is pretrained on a diverse cohort: 222 participants from nine sites, spanning clinical monitoring and cognitive tasks. PHASE-T generalizes to institutions it never saw: without local labels, it identifies pathological channels better than supervised models, and with a few labeled participants, exceeds the best published F1 by up to 83\% (Table~\ref{tab:external-transfer}).
\item \textbf{PHASE achieves strong performance across clinical and cognitive applications.} On all Omni-iEEG clinical tasks~\citep{duan2026omniieeg}, spanning brief events, brain states, and regional anatomy, PHASE-T outperforms every evaluated foundation model when frozen, with margins averaging 10.4\% across the seven scores and up to 31.1\% in anatomical localization, and sets a new state of the art when fine-tuned (Table~\ref{tab:omni}), exceeding the performance of task-specific supervised models. In naturalistic listening on BrainTreebank~\citep{wang2024brain}, frozen PHASE-ST decodes volume and pitch better than published models (Table~\ref{tab:braintreebank-auditory}).

\item \textbf{PHASE encapsulates known physiological characteristics.} Although PHASE-T is pretrained without ictal recordings, electrode coordinates, or anatomical labels, retrospective analyses show that known physiological
characteristics, including spatiotemporal patterns associated
with seizure onset and propagation and regional anatomical
organization, can be observed in the learned embedding space,
with anatomical correspondence extending across institutions (Figure~\ref{fig:readings}). PHASE-ST draws on the synchrony of a participant's concurrently recorded channels to identify the seizure-onset zone (SOZ), with relation targets accounting for most of the gain from concurrent context (Figure~\ref{fig:phasest}).
\end{enumerate}

\section{Related Work}
\label{sec:related-work}

\textbf{Pretraining objectives.}
Pretraining targets determine which signal properties receive
explicit supervision.
BrainBERT reconstructs masked
spectrograms~\citep{wang2023brainbert}, and Brant reconstructs waveform
patches from temporal and band-power inputs~\citep{zhang2023brant}.
MVPFormer instead contrastively predicts future wavelet
embeddings~\citep{carzaniga2026mvpformer}.
Beyond the prediction objective, preprocessing determines which
aspects of signal scale remain available for reconstruction.
BrainBERT z-scores spectrograms within each frequency bin,
emphasizing relative temporal variation rather than absolute
spectral magnitude.
Reconstructing these targets therefore does not require recovering
the spectral scale removed by normalization.
PHASE-T follows the joint-embedding predictive architecture (JEPA)
approach~\citep{assran2023jepa} and pairs latent prediction with a compact
set of targets measured from the signal: multi-scale wavelet power,
amplitude and local variation, and short-window high-frequency power.
These targets are standardized using fixed training-set statistics,
preserving a common reference across recordings.

\textbf{Temporal coverage.}
Learning broadly useful iEEG representations requires both fine
temporal detail and extended context.
Existing models capture waveform structure and temporal dependencies
at different scales: BrainBERT~\citep{wang2023brainbert} and
BaRISTA~\citep{oganesian2025barista} encode 5-s and 3-s frames, whereas
Brant models dependencies between 6-s patches~\citep{zhang2023brant}.
These configurations resolve either fine structure within short windows
or longer dependencies between coarse patches, but not both across a
full minute.
PHASE combines these two scales through a fine--coarse--fine
hierarchy that produces dense, fine-resolution features over
60-s inputs.

\textbf{Channel organization.}
Temporal learning and channel composition can be coupled within a model or separated into pretrained stages. Brant applies temporal and spatial attention
sequentially~\citep{zhang2023brant}.
BaRISTA structures encoding and masking by anatomical
scales~\citep{oganesian2025barista}; NeuroCLUS learns functional
channel groupings~\citep{zheng2026neuroclus}.
Closest to our design, PopT composes frozen temporal features with a coordinate-aware population
Transformer~\citep{chau2025popt}.
PHASE-ST also composes frozen temporal features, but needs no electrode
coordinates and supervises the composition with measured cross-channel
relationships (phase-locking value, envelope and broadband correlation, and
signed lag) in addition to latent prediction.

\textbf{Clinical and cognitive evaluation.}
Evidence for a shared backbone must span different prediction scales,
functional demands, and recording populations.
Task-specific models illustrate these varied requirements through
HFO morphology~\citep{zhang2025pathologicalhfo},
fast-ripple learning from detector agreement~\citep{zhang2025ss2ld},
channel and regional interactions~\citep{chen2022brainnet}, and
speech decoding from electrocorticography~\citep[ECoG;][]{makin2020translation}.
Prior iEEG foundation models demonstrate complementary capabilities:
BrainBERT is evaluated on auditory decoding~\citep{wang2023brainbert},
whereas Brant is evaluated on forecasting, imputation, and seizure
detection~\citep{zhang2023brant}.
MVPFormer combines auditory decoding with clinical transfer,
but its clinical evaluation centers on seizure
detection~\citep{carzaniga2026mvpformer}.
These evaluations do not establish whether a single frozen backbone
supports the full range of prediction levels from transient-event
classification through brain-state and channel characterization
to participant-level outcomes, together with cross-institutional transfer.
We evaluate one pretrained PHASE-T across these levels on Omni-iEEG, together with cross-institutional transfer and auditory decoding (Section~\ref{evaluation-tasks}).

\section{Method}\label{sec:method}
\label{physically-grounded-representation-learning}

\begin{figure}[t]
\centering
\includegraphics[width=0.85\textwidth]{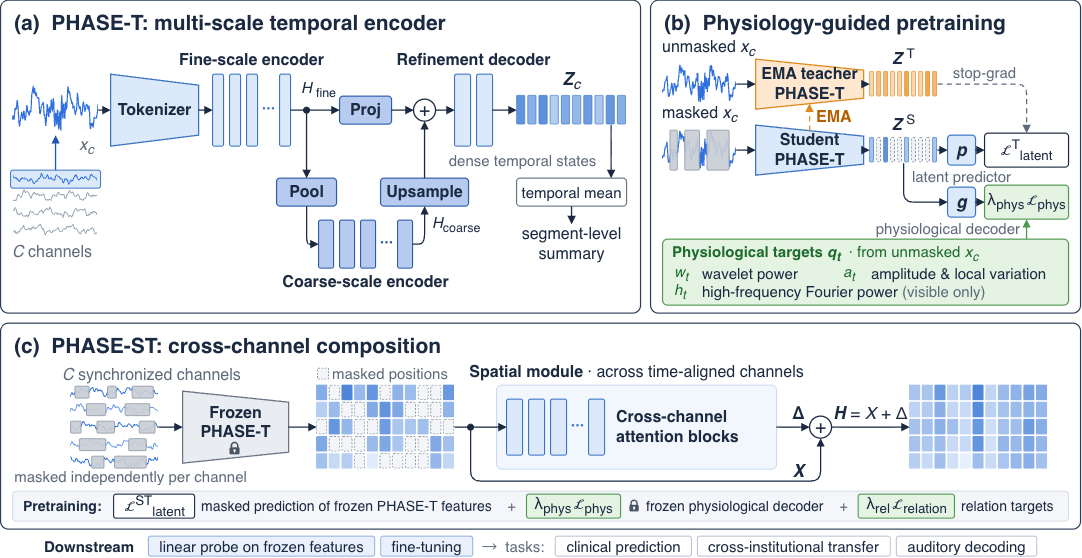}
\caption{\textbf{PHASE separates shared temporal encoding from cross-channel
composition.}
\textbf{a}, PHASE-T, the temporal encoder shared across channels, whose
refinement decoder combines fine- and coarse-scale features into
temporal states $Z_c$.
\textbf{b}, Pretraining with masked latent prediction and physiological
targets.
\textbf{c}, PHASE-ST, which updates frozen PHASE-T features across synchronized
channels; rows are channels and columns are time steps.
The bottom row shows downstream use.}
\label{fig:phase-overview}
\end{figure}

PHASE is pretrained in two stages (Figure~\ref{fig:phase-overview}).
PHASE-T first encodes each channel independently, connecting local waveform
structure with context spanning up to one minute.
PHASE-ST then uses synchronized channels to update these representations
while keeping the temporal encoder frozen.
Both stages pair latent prediction with explicit targets for physiological
characteristics measured from individual waveforms; PHASE-ST adds targets
for measured relationships between channels.

\subsection{A Shared Temporal Representation across Scales}
\label{model-architecture}
\label{single-channel-encoder}

PHASE-T brings brief waveform structure and sustained neural context into one dense temporal representation. Sharing the encoder across channels establishes a common feature space without electrode coordinates or fixed channel identities. The fine--coarse--fine hierarchy connects these temporal scales (Figure~\ref{fig:phase-overview}a). For the waveform $x_c$ of channel $c$, the encoder produces
$Z_c=f_\theta(x_c)\in\mathbb R^{N\times d}$ on a 32-ms grid,
where $N$ is the number of temporal tokens and $d=768$ is the feature dimension.
Shifted local attention encodes overlapping waveform patches into
fine-scale features $H_{\mathrm{fine}}$, while global attention over a
pooled, coarser sequence spans inputs of up to one minute, yielding coarse
contextual features $H_{\mathrm{coarse}}$.
The refinement decoder combines upsampled context with local features
through a skip connection:
\begin{equation}
\label{eq:refinement-decoder}
Z_c=
\operatorname{Decoder}\!\left(
\operatorname{Upsample}(H_{\mathrm{coarse}})
+\operatorname{Proj}(H_{\mathrm{fine}})
\right)
\in\mathbb R^{N\times d},
\end{equation}
where $\operatorname{Upsample}$ learns to map coarse features back to
the original token grid, $\operatorname{Proj}$ linearly projects local
features to width $d$, and $\operatorname{Decoder}$ refines their sum.
The final states serve as time-resolved features, and their temporal mean
provides a summary for segment-level prediction.

\subsection{Grounding Latent Prediction in Electrophysiology}
\label{single-channel-pretraining}

PHASE-T combines two complementary objectives on the same dense temporal states (Figure~\ref{fig:phase-overview}b).
Latent prediction learns dependencies across masked spans, while physiological
supervision anchors these states to properties measured from the signal.
For latent prediction, we follow JEPA~\citep{assran2023jepa}: a student processes temporally masked inputs, and a
predictor estimates the states of its unmasked EMA teacher.
Mean squared error between feature-normalized predictions and teacher targets
at masked positions defines $\mathcal L_{\mathrm{latent}}^{\mathrm T}$.

A physiological decoder supervises the same student states with
electrophysiological properties measured directly from the signal.
It predicts
$q_t=[w_t;a_t;h_t]$ at each token position $t$, where
$w_t$ contains multi-scale wavelet power,
$a_t$ amplitude and local-variation measures, and
$h_t$ short-window high-frequency Fourier power.
These targets are measured from unmasked waveforms and standardized using fixed training statistics shared across recordings. Wavelet and amplitude targets supervise masked prediction positions and fully
visible positions. Local Fourier targets supervise fully visible patches only.
Frequency-dependent targets contribute only within the recording's effective
bandwidth.
The physiological loss $\mathcal L_{\mathrm{phys}}$ averages Smooth L1 errors
within each target group and then equally across active groups.
With $\lambda_{\mathrm{phys}}$ weighting the physiological term,
the student encoder, latent predictor, and physiological decoder are jointly
optimized with
\begin{equation}
\label{eq:temporal-objective}
\mathcal L_{\mathrm{temporal}}
=\mathcal L_{\mathrm{latent}}^{\mathrm T}
+\lambda_{\mathrm{phys}}\mathcal L_{\mathrm{phys}}.
\end{equation}

\subsection{Composing Temporal Features across Channels}
\label{multichannel-pretraining}
\label{multichannel-encoder}

PHASE-ST represents each channel in the context of the participant's simultaneously
recorded channels, without electrode coordinates and for montages of any size.
It takes the frozen PHASE-T features of $C$ synchronized channels,
$X\in\mathbb R^{C\times N\times d}$, and applies self-attention across channels
independently at each time position.
Its Transformer blocks produce a residual update $\Delta$ in the temporal feature
space, giving spatial states $H=X+\Delta$ (Figure~\ref{fig:phase-overview}c).
Parameters are shared across channels and no channel identities or coordinates
enter the model, so $H$ is permutation equivariant and one set of parameters
serves any number of channels.

PHASE-ST is pretrained with three objectives: latent prediction and
physiological supervision on the spatial states, and relation supervision of
channel pairs:
\begin{equation}
\label{eq:spatial-objective}
\mathcal L_{\mathrm{spatial}}
=\mathcal L_{\mathrm{latent}}^{\mathrm{ST}}
+\lambda_{\mathrm{phys}}\mathcal L_{\mathrm{phys}}
+\lambda_{\mathrm{rel}}\mathcal L_{\mathrm{relation}}.
\end{equation}
For latent prediction, waveform spans are masked independently in each channel,
and PHASE-ST is trained to predict the clean, frozen PHASE-T features at the
masked positions, so the activity missing from one channel must be inferred from
its own context and its synchronized peers.
Physiological supervision applies PHASE-T's frozen physiological decoder to the
spatial states with the targets defined above, so the states
stay anchored to the signal properties, including at masked positions.

Relation supervision extends physiological guidance from single channels to
their interactions.
From the clean waveforms, we measure phase-locking value~\citep{lachaux1999measuring}, envelope correlation~\citep{bruns2000amplitude},
broadband correlation, and signed lag between channel pairs, which characterize
how activity couples and propagates across channels; relation heads predict them
from temporal averages of the spatial states.
Predicted phase locking and correlations do not depend on channel order, while
lag changes sign when the channels are exchanged.
PHASE-T stays frozen: only the spatial module, latent predictor, and relation
heads are trained.
Implementation details for both stages are given in
Appendices~\ref{app:model-pretraining-details} and~\ref{app:channel-set-method-details}.

\section{Experimental Setup}\label{experimental-setup}

\subsection{Pretraining Data and Training Stages}
\label{pretraining-data-and-stages}

The pretraining corpus spans two recording contexts: clinical monitoring from
Omni-iEEG~\citep{duan2026omniieeg} and Hatano Detroit~\citep{hatano2026validation},
and naturalistic listening from NYU Podcast ECoG~\citep{zada2025podcast} and
BrainTreebank~\citep{wang2024brain}.
Within each source, sampling is balanced across participants.
No downstream evaluation data enter pretraining: Omni-iEEG test participants, the Tohoku, NCNP, MAYO, and FNUSA cohorts, and BrainTreebank's validation and test recordings are all excluded.

Both pretraining stages use 60-second inputs at 1\,kHz.
PHASE-T uses a 106.7M-parameter encoder and is evaluated after 60,000
pretraining updates (global batch of 192 crops; 2.8 GPU-days on four
NVIDIA PRO 6000 GPUs).
We then freeze PHASE-T and pretrain PHASE-ST on synchronized 64-channel
sets from the same corpus. PHASE-ST adds a spatial module with 3.0M
trainable parameters and is evaluated after 5,000 pretraining updates
(global batch of 128 sets; approximately 1.9 GPU-days on two
NVIDIA PRO 6000 GPUs).
Source composition, data partitions, and full pretraining settings are detailed in
Appendices~\ref{app:data-signal-preparation}--\ref{app:channel-set-method-details}.

\subsection{Downstream Evaluation}
\label{evaluation-tasks}

We evaluate PHASE-T on the five clinical tasks of
Omni-iEEG~\citep{duan2026omniieeg}, following its official task definitions,
splits, evaluation units, and metrics.
The five tasks, with the abbreviations used in Table~\ref{tab:omni}, are pathological HFO
classification (HFO), ictal period classification (Ictal), sleep--awake
classification (Sleep), anatomical localization into five lobes (Lobe-5) or
twelve regions (Region-12), and pathological brain region identification.
Omni-iEEG evaluates this task using two criteria that a model should meet
together~\citep{duan2026omniieeg}: distinguishing SOZ channels from
non-resected channels of seizure-free participants (Channel), and predicting
post-operative seizure freedom from the resection ratio of the model's scores
(Outcome). This ratio is the share of a participant's total pathology score
that falls in resected channels.
We compare frozen PHASE-T with the released
foundation models BrainBERT~\citep{wang2023brainbert},
BaRISTA~\citep{oganesian2025barista}, and
MVPFormer~\citep{carzaniga2026mvpformer}, using linear probes.
We compare fine-tuned PHASE-T with the benchmark's published models
PyHFO-Omni~\citep{zhang2024pyhfo}, CLAP~\citep{laionclap2023}, TimeConv-CNN~\citep{duan2026omniieeg},
and SEEG-NET~\citep{wang2022seegnet}.

We assess cross-institutional transfer by applying the Omni-iEEG-trained
pathological-channel probe unchanged to Hatano's Tohoku and NCNP
cohorts~\citep{hatano2026validation}. Following Omni-iEEG, we train CLAP and
SEEG-NET on the same Omni-iEEG task and apply them without adaptation.
We further evaluate cross-participant pathology classification at MAYO and
FNUSA~\citep{nejedly2020multicenter}, following MVPFormer's data split:
we fine-tune PHASE-T on four participants at each institution
and test it on the remaining participants.
We compare its scores with the MVPFormer and NeuroCLUS results reported
by \citet{zheng2026neuroclus}.

For multichannel auditory decoding on BrainTreebank~\citep{wang2024brain},
we evaluate both frozen and fine-tuned PHASE-ST against the published results
of PopT (with BrainBERT features), BaRISTA, MVPFormer, and NeuroCLUS,
following BaRISTA's recording split and
PopT's evaluation protocol~\citep{chau2025popt}.
Literature results are reported at their published precision.
Implementation details of downstream evaluation are provided in
Appendix~\ref{app:downstream-implementation}.

\section{Results}\label{results}

\subsection{Clinical Reach across Temporal Scales and Institutions}
\label{clinical-transfer-across-scales}

On the Omni-iEEG test set, PHASE-T handles all five clinical tasks with linear probes on the same frozen model, from HFO events lasting tens of milliseconds to ictal periods and sleep--awake states in one-minute windows, and the anatomical origin of a channel (Table~\ref{tab:omni}).
In the frozen single-channel comparison, it leads three foundation-model baselines on every task, by the widest margins on Region-12, Lobe-5, and Ictal, which depend on regional physiology and minute-scale context.
Its Region-12 macro-F1 is 31\% higher than that of BrainBERT, the strongest of the three.
Frozen PHASE-T outperforms specialist models on HFO, Sleep, Lobe-5, and Channel, although each is trained for particular tasks with fixed input formats and prediction targets.
Fine-tuned PHASE-T sets a new state of the art on all tasks, leading on both Omni-iEEG's pathology criteria: its Outcome AUC of 0.7713 exceeds that of PyHFO-Omni, an event-based model built on expert HFO annotations.

\begin{table*}[t]
\centering
\footnotesize
\setlength{\tabcolsep}{3pt}
\caption{\textbf{One representation across five clinical tasks, from HFO events to surgical outcome.}
Results are on the official Omni-iEEG test participants. Frozen PHASE-T and the three foundation models are evaluated with linear probes; fine-tuned PHASE-T is compared with the specialist models published by Omni-iEEG.
$C{=}1$ marks multichannel models run on one channel at a time. F1 is macro-F1 and AUC is ROC-AUC. Dashes denote tasks that a specialist model does not support.
Bold marks the highest performance per column, and underline the highest among frozen encoders.}
\label{tab:omni}

\begin{tabular}{@{}lrrrrrrr@{}}
\toprule
 & Event & \multicolumn{2}{c}{Window (60 s)} & \multicolumn{2}{c}{Anatomy} & \multicolumn{2}{c}{Pathol. brain region} \\
\cmidrule(lr){2-2}\cmidrule(lr){3-4}\cmidrule(lr){5-6}\cmidrule(lr){7-8}
Model & HFO & Ictal & Sleep & Lobe-5 & Region-12 & Channel & Outcome \\
 & F1 & F1 & F1 & F1 & F1 & AUC & AUC \\
\midrule
PyHFO-Omni
  & 0.8061 & -- & -- & -- & -- & 0.7351 & 0.7438 \\
CLAP
  & -- & 0.9245 & 0.7225 & 0.4750 & 0.3540 & 0.7684 & 0.6770 \\
TimeConv-CNN
  & -- & 0.8533 & 0.7118 & 0.4788 & 0.3087 & 0.8061 & 0.7380 \\
SEEG-NET
  & -- & 0.7526 & 0.6773 & 0.2520 & 0.1081 & 0.7850 & 0.5952 \\
\midrule
BrainBERT
  & 0.7937 & 0.8323 & 0.6941 & 0.4054 & 0.2592 & 0.8055 & 0.6017 \\
BaRISTA ($C{=}1$)
  & 0.6050 & 0.7428 & 0.7213 & 0.3006 & 0.1658 & 0.6949 & 0.5739 \\
MVPFormer ($C{=}1$)
  & 0.4347 & 0.8311 & 0.5404 & 0.2423 & 0.1069 & 0.6604 & 0.5878 \\
\addlinespace[2pt]
PHASE-T, frozen
  & \underline{0.8138} & \underline{0.8958} & \underline{0.7413} & \underline{0.4822} & \underline{0.3398} & \underline{0.8394} & \underline{0.6369} \\
PHASE-T, fine-tuned
  & \textbf{0.8188} & \textbf{0.9327} & \textbf{0.8150} & \textbf{0.5134} & \textbf{0.3572} & \textbf{0.8502} & \textbf{0.7713} \\
\bottomrule
\end{tabular}%
\end{table*}

\textbf{Physiological supervision and fine-scale feature
fusion improve downstream performance.}
\label{temporal-ablation-results}
We compare full PHASE-T with three pretraining variants (Table~\ref{tab:mechanism-ablations}): Latent-only drops the physiological targets, Latent+Recon replaces them with Brant-style reconstruction~\citep{zhang2023brant} of each masked token's clean 128-sample waveform patch, and No-skip removes the fine-stage skip connection.
All four models are frozen after 60,000 pretraining updates and probed with the same protocol.
Over Latent-only, physiological supervision adds the most macro-F1 on brief events (HFO, 0.63 to 0.81), sleep (0.54 to 0.74), and anatomical localization (Region-12, 0.20 to 0.34).
Latent+Recon leaves HFO and Region-12 at the Latent-only level, and full PHASE-T outperforms it on all five tasks, so physiological supervision is a more useful pretraining signal for these tasks than sample-level waveform reconstruction.
The fine-stage skip connection adds 0.15 macro-F1 on sleep and 0.13 on ictal over No-skip, showing that fine-scale features complement the coarse pathway's context.

\begin{table}[t]
\centering
\footnotesize
\setlength{\tabcolsep}{3pt}
\renewcommand{\arraystretch}{1.08}
\caption{\textbf{Ablation Studies.}
Full PHASE-T is the unablated model, and each variant changes one component. All rows are evaluated under the same pretraining and linear probe conditions; the Full PHASE-T row repeats the frozen scores of Table~\ref{tab:omni}.
F1 is macro-F1 and AUC is ROC-AUC.
Bold marks the highest performance per column.}
\label{tab:mechanism-ablations}
\begin{tabular}{@{}lrrrrrrr@{}}
\toprule
 & Event & \multicolumn{2}{c}{Window (60 s)} & \multicolumn{2}{c}{Anatomy} & \multicolumn{2}{c}{Pathol. brain region} \\
\cmidrule(lr){2-2}\cmidrule(lr){3-4}\cmidrule(lr){5-6}\cmidrule(lr){7-8}
Pretraining variant & HFO & Ictal & Sleep & Lobe-5 & Region-12 & Channel & Outcome \\
 & F1 & F1 & F1 & F1 & F1 & AUC & AUC \\
\midrule
Full PHASE-T
  & \textbf{0.8138} & \textbf{0.8958} & \textbf{0.7413}
  & \textbf{0.4822} & \textbf{0.3398} & \textbf{0.8394} & \textbf{0.6369} \\
Latent-only
  & 0.6276 & 0.8597 & 0.5398 & 0.3271 & 0.1996 & 0.7570 & 0.6285 \\
Latent+Recon
  & 0.6298 & 0.8532 & 0.6405 & 0.3126 & 0.1954 & 0.7745 & 0.6156 \\
No-skip
  & 0.7093 & 0.7684 & 0.5904 & 0.3640 & 0.2371 & 0.7415 & 0.6354 \\
\bottomrule
\end{tabular}
\end{table}

\textbf{PHASE-T transfers to institutions held out from pretraining.}
\label{cross-institution-transfer-results}
Generalization to new institutions is important for the clinical utility of pathology models, as recording setups differ and expert labels are scarce.
Fitted only on Omni-iEEG, the PHASE-T probe identifies pathological channels at Tohoku and NCNP better than every supervised model trained on the same data, most clearly at NCNP (ROC-AUC 0.6292 against 0.5528; Table~\ref{tab:external-transfer}).
For cross-participant pathology classification, fine-tuned PHASE-T nearly doubles the best published F1 at MAYO (0.7337 against 0.40; \citealt{zheng2026neuroclus}) and exceeds it at FNUSA.

\begin{table*}[t]
\centering
\hypersetup{hidelinks}
\footnotesize
\setlength{\tabcolsep}{3pt}
\caption{\textbf{Cross-institution generalization with no local labels or labels from four participants.}
\textbf{(a)} Pathological-channel identification at Tohoku and NCNP: the PHASE-T probe and supervised comparators are trained on Omni-iEEG and applied unchanged.
\textbf{(b)} Pathology classification at MAYO and FNUSA after fine-tuning on four local participants; comparator values are published results.
Bold marks the highest performance per column.}
\label{tab:external-transfer}

\begin{minipage}[t]{0.48\textwidth}
\vspace{0pt}
\textbf{(a) Cross-institution transfer (ROC-AUC)}\strut\par
\vspace{4pt}
\begin{tabular*}{\linewidth}{@{\extracolsep{\fill}}lrr@{}}
\toprule
Model & Tohoku & NCNP \\
\midrule
CLAP & 0.7663 & 0.5021 \\
SEEG-NET & 0.7636 & 0.5528 \\
\midrule
PHASE-T, frozen & \textbf{0.7775} & \textbf{0.6292} \\
\bottomrule
\end{tabular*}
\end{minipage}\hfill
\begin{minipage}[t]{0.48\textwidth}
\vspace{0pt}
\textbf{(b) Cross-participant fine-tuning (binary F1)}\strut\par
\vspace{4pt}
\begin{tabular*}{\linewidth}{@{\extracolsep{\fill}}lrr@{}}
\toprule
Model & MAYO & FNUSA \\
\midrule
MVPFormer & 0.36\phantom{00} & 0.46\phantom{00} \\
NeuroCLUS & 0.40\phantom{00} & 0.51\phantom{00} \\
\midrule
PHASE-T, fine-tuned & \textbf{0.7337} & \textbf{0.6656} \\
\bottomrule
\end{tabular*}
\end{minipage}
\end{table*}

\subsection{Label-Free Seizure Dynamics and Regional Identity}
\label{sec:readings}

\begin{figure}[t]
\centering
\includegraphics[width=0.85\textwidth]{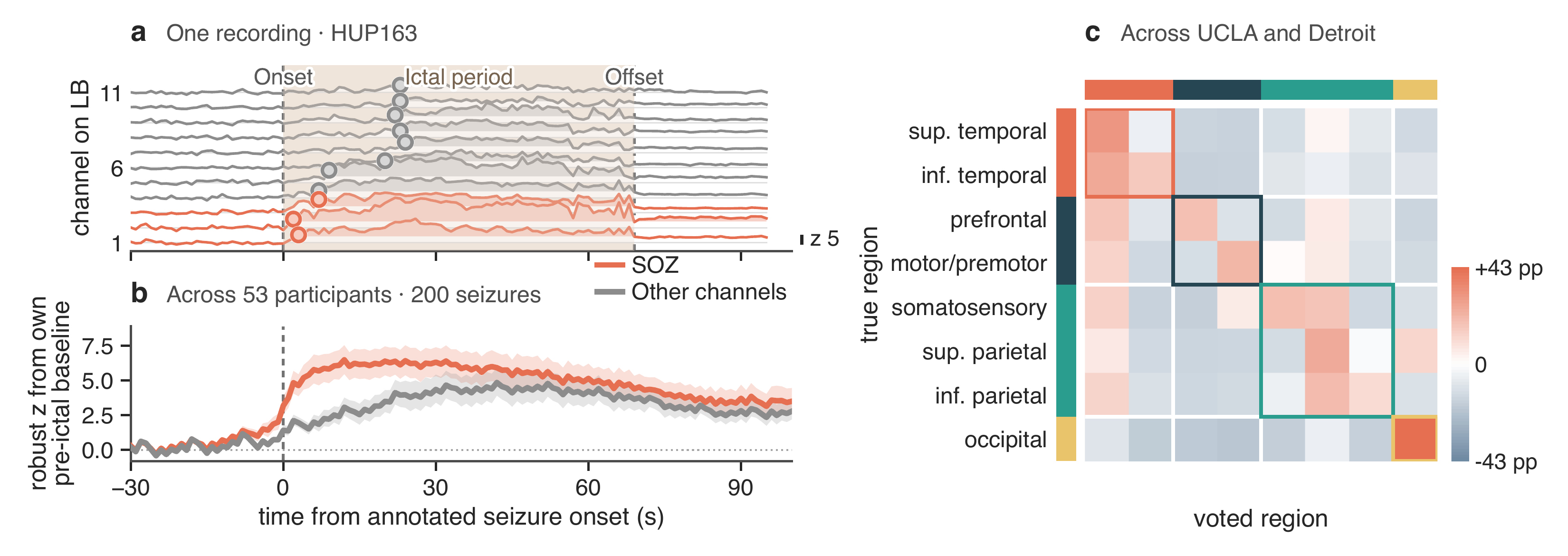}
\caption{\textbf{Frozen PHASE-T embeddings carry seizure dynamics and regional
identity across institutions.}
\textbf{a}, Embedding distance from the pre-ictal baseline for each channel on
one stereo-EEG shaft before, during, and after one ictal period; orange traces are SOZ
channels, and circles mark departures.
\textbf{b}, The same distance for SOZ and other channels: the per-seizure
median over channels, averaged over each participant's seizures and then over
participants; shading shows 95\% participant-bootstrap confidence intervals.
\textbf{c}, The proportion of each true region's channels (rows) voted into
each region (columns), pooled over both directions between UCLA and Detroit
and shown relative to chance (red, above; blue, below). Each channel is voted on
by its ten most similar participants at the other institution; boxes group
regions by lobe.}
\label{fig:readings}
\end{figure}

Task scores measure how useful PHASE-T's embeddings are, not what they encode.
We therefore analyze the frozen embeddings directly, without updating the encoder or fitting a linear probe, and find that they encode two kinds of structure PHASE-T was never explicitly taught: how seizures unfold, and what distinguishes brain regions across institutions (Figure~\ref{fig:readings}).

\textbf{PHASE-T captures how seizures unfold across space and time.}
We analyze 200 seizures from 53 participants recorded at the Hospital of the University of Pennsylvania~\citep[HUP;][]{bernabei2023quantitative} within Omni-iEEG, using the release's annotations of ictal period and SOZ channels.
For each channel, we measure the distance of its frozen embedding from its own pre-ictal baseline (up to 90~s ending 10~s before onset) in a space of up to 32 whitened principal components, and express this distance as a robust $z$-score against the baseline; a channel departs when it first stays above $z=5$ for five consecutive seconds.
When an ictal period begins, channels in the clinically marked seizure-onset zone move farther from baseline and peak earlier than the rest (Figure~\ref{fig:readings}b), although PHASE-T's pretraining data contain no ictal recordings.
PHASE-T encodes each channel independently and receives no electrode positions, yet departure times derived from its embeddings are ordered along the array: channels farther from the first departing channel depart later (Figure~\ref{fig:readings}a; median Spearman correlation 0.60 over 706 stereo-EEG shaft--seizure pairs at HUP and 0.66 over 75 ECoG strip--seizure pairs from three other institutions~\citep{li2021neural}; 0.34 and 0.39 with departure times shuffled within each array).
PHASE-T captures the distinct dynamics of SOZ channels and how seizures unfold along the electrode array, reflecting the spatiotemporal organization of human seizures~\citep{martinet2017slow}.

\textbf{PHASE-T learns a map of the cortex that holds across institutions.}
Averaged by brain region, PHASE-T's frozen embeddings of 16,444 ECoG channels in the Omni-iEEG anatomy recordings line up across institutions: all eight Region-12 regions we compare at UCLA (32 participants) are most similar to their own counterparts at Detroit (135 participants), and six of the eight Detroit regions to theirs at UCLA.
These 14 matches out of 16 are far above chance: about seven times the 2 expected at random, and a count that none of 2,000 shuffles of the region labels reached (permutation test, $p=5\times10^{-4}$).
The different regions whose averages lie closest together are neighbors in the brain: motor and somatosensory cortex, parietal and occipital cortex, and the two temporal regions (Appendix~\ref{app:anatomical-correspondence}).
The map reaches single channels: voted on by their ten most similar participants at the other institution, the channels of every region land in their own region more often than chance, most often in their own region or, for inferior temporal and inferior parietal cortex, in the superior region of the same lobe (Figure~\ref{fig:readings}c).
PHASE-T's lead in anatomical localization (Table~\ref{tab:omni}) thus reflects a map of the cortex that its frozen embeddings carry consistently across institutions.

\subsection{Context from the Concurrent Montage}
\label{sec:phasest}

\begin{figure}[t]
\centering
\includegraphics[width=0.85\textwidth]{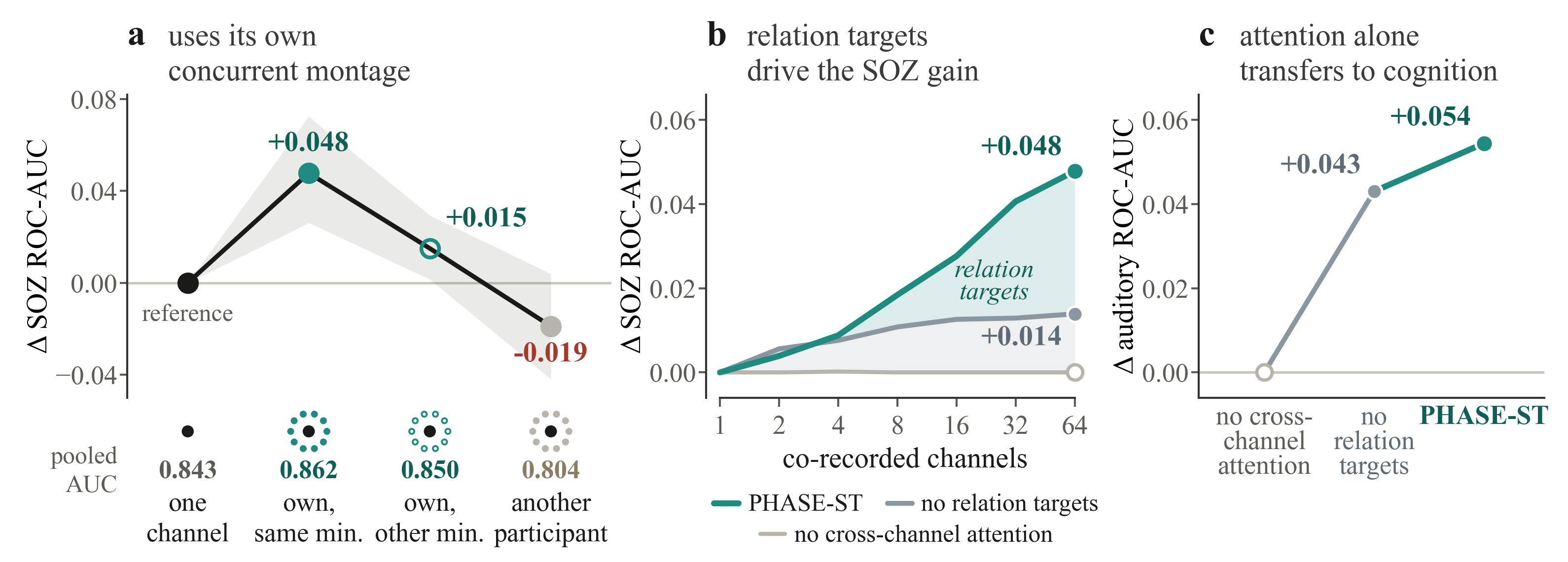}
\caption{\textbf{PHASE-ST uses a participant's concurrent montage, learns this
from relation targets, and carries context into auditory decoding.}
\textbf{a}, Per-participant mean change in SOZ ROC-AUC relative to scoring the
target channel alone, for each set of companion channels; shading shows the
95\% participant-bootstrap confidence interval. Glyphs depict the target channel
(black) and its companions, with each condition's pooled ROC-AUC below.
\textbf{b}, The same change as a function of the number of co-recorded
channels.
\textbf{c}, Change in auditory ROC-AUC on BrainTreebank relative to the stage
without cross-channel attention, averaged over four tasks and seven participants.}
\label{fig:phasest}
\end{figure}

PHASE-ST interprets each channel in the context of the other channels recorded at the same time.
In naturalistic listening on BrainTreebank, its frozen features decode volume and pitch better than published models (Table~\ref{tab:braintreebank-auditory}).
In clinical recordings, it draws on the synchrony of a participant's concurrently recorded channels to identify SOZ channels more accurately than PHASE-T; relation targets account for most of the gain from this concurrent context (Figure~\ref{fig:phasest}).

\textbf{PHASE-ST decodes acoustic properties during naturalistic listening.}
\label{auditory-decoding-results}
We evaluate PHASE-ST on four auditory decoding tasks across seven BrainTreebank participants (Table~\ref{tab:braintreebank-auditory}).
Frozen PHASE-ST reaches mean ROC-AUCs of 0.9450 for volume and 0.8822 for pitch, above the strongest published results of 0.92 and 0.83~\citep{zheng2026neuroclus}.
Fine-tuning raises the scores on all four tasks (volume to 0.9515, pitch to 0.8886).

\begin{table*}[t]
\centering
\footnotesize
\caption{\textbf{Auditory decoding on BrainTreebank.}
Results are mean test ROC-AUC across the seven evaluation participants. Literature rows are published results.
Bold marks the highest and underline the second-highest performance per column.}
\label{tab:braintreebank-auditory}
\begin{tabular*}{\textwidth}{@{\extracolsep{\fill}}lcccc@{}}
\toprule
Method & Sentence onset & Speech & Volume & Pitch \\
\midrule
PopT + BrainBERT & 0.90 & 0.93 & 0.87 & 0.74 \\
BaRISTA & 0.91 & 0.92 & 0.85 & 0.73 \\
MVPFormer & 0.87 & 0.90 & 0.88 & 0.83 \\
NeuroCLUS & \textbf{0.96} & \textbf{0.99} & 0.92 & 0.83 \\
\midrule
PHASE-ST, frozen &
0.9176 & 0.9703 & \underline{0.9450} & \underline{0.8822} \\
PHASE-ST, fine-tuned &
\underline{0.9245} & \underline{0.9786} & \textbf{0.9515} & \textbf{0.8886} \\
\bottomrule
\end{tabular*}
\end{table*}

\textbf{PHASE-ST draws on synchrony with concurrently recorded channels to identify the SOZ.}
With the participant's own channels from the same minute, in panels of up to 64, PHASE-ST reaches a pooled ROC-AUC of 0.862, compared with 0.843 when each channel is processed alone and 0.841 for PHASE-T on the same channels.
The added context yields a mean per-participant gain of $+0.048$ across 52 held-out Omni-iEEG participants (Figure~\ref{fig:phasest}a).
This advantage depends on synchrony with the target channel.
Taking the same companion channels from another minute preserves the participant and recording setup and keeps the companions mutually synchronized, while breaking their alignment with the unchanged target waveform.
With linear probes fitted to their respective conditions, the mean gain falls to $+0.015$, significantly below the same-minute gain (two-sided sign test, $p=0.003$).
Replacing the companions with another participant's channels instead gives a mean per-participant change of $-0.019$ under the fixed same-minute probe.

\textbf{Relation targets drive the SOZ context gain, and cross-channel attention carries context into auditory decoding.}
We compare PHASE-ST with two variants trained identically except for one component: one without relation targets and one without cross-channel attention.
PHASE-ST's SOZ context gain grows with the number of co-recorded channels to $+0.048$ (95\% CI [$+0.026$, $+0.071$]) at the largest panels of up to 64 channels; without relation targets it reaches only $+0.014$, so $+0.034$ of the gain comes from the relation targets, and without cross-channel attention it stays at zero by construction (Figure~\ref{fig:phasest}b).
On BrainTreebank, PHASE-ST improves decoding over the stage without cross-channel attention for all seven participants, by a mean $+0.054$ ROC-AUC. The variant without relation targets has a mean gain of $+0.043$ over the same stage (Figure~\ref{fig:phasest}c).
Appendix~\ref{app:st-context} details the context probe and spatial-stage ablations.

\section{Conclusion and Limitations}
\label{sec:conclusion}

PHASE uses physiology-guided pretraining to learn
temporal and cross-channel representations of iEEG.
PHASE-T leads the evaluated foundation models on all five
Omni-iEEG clinical tasks with frozen representations and
sets a new state of the art after fine-tuning, while
PHASE-ST achieves the best volume and pitch decoding
among compared methods.
PHASE-T's embeddings also capture seizure dynamics and
consistent regional organization across institutions.
Together, these results support PHASE as a physiology-guided
foundation for clinical and cognitive iEEG analysis.
The framework uses separate temporal
and spatial pretraining stages and context limited to one minute.
Future work will explore joint pretraining, extend context
toward hour-long recordings, and expand both patient cohorts
and downstream evaluations to assess broader generalization
(Appendix~\ref{app:limitations}).

\bibliography{references}

@inproceedings{duan2026omniieeg,
  author    = {Duan, Chenda and Zhang, Yipeng and Kanai, Sotaro and
               Ding, Yuanyi and Daida, Atsuro and Yu, Pengyue and
               Zheng, Tiancheng and Kuroda, Naoto and Hussain, Shaun A. and
               Asano, Eishi and Nariai, Hiroki and Roychowdhury, Vwani},
  title     = {{Omni-iEEG}: A Large-Scale, Comprehensive {iEEG} Dataset and
               Benchmark for Epilepsy Research},
  booktitle = {International Conference on Learning Representations},
  year      = {2026},
  url       = {https://openreview.net/forum?id=rv9lQpY5cG}
}

@inproceedings{zhang2025ss2ld,
  author    = {Zhang, Yipeng and Ding, Yuanyi and Duan, Chenda and
               Daida, Atsuro and Nariai, Hiroki and Roychowdhury, Vwani},
  title     = {Self-Supervised Distillation of Legacy Rule-Based Methods for
               Enhanced {EEG}-Based Decision-Making},
  booktitle = {Medical Image Computing and Computer Assisted Intervention -- MICCAI 2025},
  series    = {Lecture Notes in Computer Science},
  volume    = {15967},
  pages     = {477--486},
  publisher = {Springer Nature Switzerland},
  year      = {2025},
  doi       = {10.1007/978-3-032-04984-1_46}
}

@article{zhang2022physiologicalhfo,
  author  = {Zhang, Yipeng and Chung, Hoyoung and Ngo, Jacquline P. and
             Monsoor, Tonmoy and Hussain, Shaun A. and Matsumoto, Joyce H. and
             Walshaw, Patricia D. and Fallah, Aria and Sim, Myung Shin and
             Asano, Eishi and Sankar, Raman and Staba, Richard J. and
             Engel, Jerome and Speier, William and Roychowdhury, Vwani and
             Nariai, Hiroki},
  title   = {Characterizing Physiological High-Frequency Oscillations
             Using Deep Learning},
  journal = {Journal of Neural Engineering},
  volume  = {19},
  number  = {6},
  pages   = {066027},
  year    = {2022},
  doi     = {10.1088/1741-2552/aca4fa}
}

@article{zhang2025pathologicalhfo,
  author  = {Zhang, Yipeng and Daida, Atsuro and Liu, Lawrence and
             Kuroda, Naoto and Ding, Yuanyi and Oana, Shingo and Kanai, Sotaro
             and Monsoor, Tonmoy and Duan, Chenda and Hussain, Shaun A. and
             Qiao, Joe X. and Salamon, Noriko and Fallah, Aria and
             Sim, Myung Shin and Sankar, Raman and Staba, Richard J. and
             Engel, Jerome and Asano, Eishi and Roychowdhury, Vwani and
             Nariai, Hiroki},
  title   = {Self-Supervised Data-Driven Approach Defines Pathological
             High-Frequency Oscillations in Epilepsy},
  journal = {Epilepsia},
  volume  = {66},
  number  = {11},
  pages   = {4434--4450},
  year    = {2025},
  doi     = {10.1111/epi.18545}
}

@article{monsoor2023hfo,
  author  = {Monsoor, Tonmoy and Zhang, Yipeng and Daida, Atsuro and
             Oana, Shingo and Lu, Qiujing and Hussain, Shaun A. and
             Fallah, Aria and Sankar, Raman and Staba, Richard J. and
             Speier, William and Roychowdhury, Vwani and Nariai, Hiroki},
  title   = {Optimizing Detection and Deep Learning-Based Classification of
             Pathological High-Frequency Oscillations in Epilepsy},
  journal = {Clinical Neurophysiology},
  volume  = {154},
  pages   = {129--140},
  year    = {2023},
  doi     = {10.1016/j.clinph.2023.07.012}
}

@article{zhang2024pyhfo,
  author    = {Zhang, Yipeng and Liu, Lawrence and Ding, Yuanyi and Chen, Xin
               and Monsoor, Tonmoy and Daida, Atsuro and Oana, Shingo and
               Hussain, Shaun and Sankar, Raman and Fallah, Aria and others},
  title     = {{PyHFO}: Lightweight Deep Learning-Powered End-to-End
               High-Frequency Oscillations Analysis Application},
  journal   = {Journal of Neural Engineering},
  volume    = {21},
  number    = {3},
  pages     = {036023},
  year      = {2024}
}

@article{wang2022seegnet,
  author  = {Wang, Yiping and Yang, Yanfeng and Cao, Gongpeng and
             Guo, Jinjie and Wei, Penghu and Feng, Tao and Dai, Yang and
             Huang, Jinguo and Kang, Guixia and Zhao, Guoguang},
  title   = {{SEEG-Net}: An Explainable and Deep Learning-Based Cross-Subject
             Pathological Activity Detection Method for Drug-Resistant Epilepsy},
  journal = {Computers in Biology and Medicine},
  volume  = {148},
  pages   = {105703},
  year    = {2022},
  doi     = {10.1016/j.compbiomed.2022.105703}
}

@inproceedings{chen2022brainnet,
  author    = {Chen, Junru and Yang, Yang and Yu, Tao and Fan, Yingying and
               Mo, Xiaolong and Yang, Carl},
  title     = {{BrainNet}: Epileptic Wave Detection from {SEEG} with
               Hierarchical Graph Diffusion Learning},
  booktitle = {Proceedings of the 28th ACM SIGKDD Conference on
               Knowledge Discovery and Data Mining},
  pages     = {2741--2751},
  year      = {2022},
  doi       = {10.1145/3534678.3539178}
}

@inproceedings{wang2023brainbert,
  author    = {Wang, Christopher and Subramaniam, Vighnesh and
               Yaari, Adam Uri and Kreiman, Gabriel and Katz, Boris and
               Cases, Ignacio and Barbu, Andrei},
  title     = {{BrainBERT}: Self-Supervised Representation Learning for
               Intracranial Recordings},
  booktitle = {The Eleventh International Conference on Learning Representations},
  year      = {2023},
  url       = {https://openreview.net/forum?id=xmcYx_reUn6}
}

@inproceedings{zhang2023brant,
  author    = {Zhang, Daoze and Yuan, Zhizhang and Yang, Yang and
               Chen, Junru and Wang, Jingjing and Li, Yafeng},
  title     = {{Brant}: Foundation Model for Intracranial Neural Signal},
  booktitle = {Advances in Neural Information Processing Systems},
  volume    = {36},
  pages     = {26304--26321},
  year      = {2023},
  doi       = {10.52202/075280-1144}
}

@article{wang2024brain,
  author  = {Wang, Christopher and Yaari, Adam and Singh, Aaditya and
             Subramaniam, Vighnesh and Rosenfarb, Dana and DeWitt, Jan and
             Misra, Pranav and Madsen, Joseph and Stone, Scellig and
             Kreiman, Gabriel and others},
  title   = {Brain Treebank: Large-Scale Intracranial Recordings from
             Naturalistic Language Stimuli},
  journal = {Advances in Neural Information Processing Systems},
  volume  = {37},
  pages   = {96505--96540},
  year    = {2024}
}

@article{li2021neural,
  author    = {Li, Adam and Huynh, Chester and Fitzgerald, Zachary and
               Cajigas, Iahn and Brusko, Damian and Jagid, Jonathan and
               Claudio, Angel O. and Kanner, Andres M. and Hopp, Jennifer and
               Chen, Stephanie and others},
  title     = {Neural Fragility as an {EEG} Marker of the Seizure Onset Zone},
  journal   = {Nature Neuroscience},
  volume    = {24},
  number    = {10},
  pages     = {1465--1474},
  year      = {2021},
  doi       = {10.1038/s41593-021-00901-w}
}

@inproceedings{he2022mae,
  title     = {Masked Autoencoders Are Scalable Vision Learners},
  author    = {He, Kaiming and Chen, Xinlei and Xie, Saining and Li, Yanghao and
               Doll{\'a}r, Piotr and Girshick, Ross},
  booktitle = {Proceedings of the IEEE/CVF Conference on Computer Vision
               and Pattern Recognition},
  pages     = {16000--16009},
  year      = {2022}
}

@inproceedings{assran2023jepa,
  title     = {Self-Supervised Learning from Images with a Joint-Embedding Predictive Architecture},
  author    = {Assran, Mahmoud and Duval, Quentin and Misra, Ishan and
               Bojanowski, Piotr and Vincent, Pascal and Rabbat, Michael and
               LeCun, Yann and Ballas, Nicolas},
  booktitle = {Proceedings of the IEEE/CVF Conference on Computer Vision
               and Pattern Recognition},
  pages     = {15619--15629},
  year      = {2023}
}

@inproceedings{oganesian2025barista,
  author    = {Oganesian, Lucine L. and Hashemi, Saba and Shanechi, Maryam M.},
  title     = {{BaRISTA}: Brain Scale Informed Spatiotemporal Representation of
               Human Intracranial Neural Activity},
  booktitle = {Advances in Neural Information Processing Systems},
  year      = {2025},
  eprint    = {2512.12135},
  archivePrefix = {arXiv},
  primaryClass  = {cs.LG}
}

@article{frauscher2018atlas,
  author  = {Frauscher, Birgit and von Ellenrieder, Nicolas and Zelmann, Rina
             and Dole\v{z}alov{\'a}, Irena and Minotti, Lorella and
             Olivier, Andr{\'e} and Hall, Jeffery and Hoffmann, Dominique
             and Nguyen, Dang Khoa and Kahane, Philippe and
             Dubeau, Fran{\c{c}}ois and Gotman, Jean},
  title   = {Atlas of the Normal Intracranial Electroencephalogram:
             Neurophysiological Awake Activity in Different Cortical Areas},
  journal = {Brain},
  volume  = {141},
  number  = {4},
  pages   = {1130--1144},
  year    = {2018},
  doi     = {10.1093/brain/awy035}
}

@article{frauscher2024ieeg,
  title={Learn how to interpret and use intracranial EEG findings},
  author={Frauscher, B and Mansilla, D and Abdallah, C and Astner-Rohracher, A and Beniczky, S and Br{\'a}zdil, Milan and Gnatkovsky, V and Jacobs, J and Kalamangalam, G and Perucca, P and others},
  journal={Epileptic Disorders},
  volume={26},
  number={1},
  pages={1--59},
  year={2024},
  publisher={Wiley Online Library},
  doi       = {10.1002/epd2.20190}
}

@article{kambara2018presurgical,
  title={Presurgical language mapping using event-related high-gamma activity: The Detroit procedure},
  author={Kambara, Toshimune and Sood, Sandeep and Alqatan, Zahraa and Klingert, Christine and Ratnam, Diksha and Hayakawa, Akane and Nakai, Yasuo and Luat, Aimee F and Agarwal, Rajkumar and Rothermel, Robert and others},
  journal={Clinical Neurophysiology},
  volume={129},
  number={1},
  pages={145--154},
  year={2018},
  publisher={Elsevier}
}

@article{andrillon2011sleep,
  title={Sleep spindles in humans: insights from intracranial EEG and unit recordings},
  author={Andrillon, Thomas and Nir, Yuval and Staba, Richard J and Ferrarelli, Fabio and Cirelli, Chiara and Tononi, Giulio and Fried, Itzhak},
  journal={Journal of Neuroscience},
  volume={31},
  number={49},
  pages={17821--17834},
  year={2011},
  publisher={Society for Neuroscience}
}

@inproceedings{
zheng2026neuroclus,
title={Neuro{CLUS}: A Foundation Model with Functional Clustering for Intracranial Neural Decoding},
author={Hui Zheng and Haiteng Wang},
booktitle={Forty-third International Conference on Machine Learning},
year={2026},
  url       = {https://openreview.net/forum?id=pFweJM4Uw8}
}

@inproceedings{chau2025popt,
  title     = {Population Transformer: Learning Population-level Representations of Neural Activity},
  author    = {Chau, Geeling and Wang, Christopher and Talukder, Sabera J and
               Subramaniam, Vighnesh and Soedarmadji, Saraswati and Yue, Yisong and
               Katz, Boris and Barbu, Andrei},
  booktitle = {The Thirteenth International Conference on Learning Representations},
  year      = {2025},
  url       = {https://openreview.net/forum?id=FVuqJt3c4L}
}

@inproceedings{carzaniga2026mvpformer,
  title     = {A Foundation Model with Multi-Variate Parallel Attention to Generate Neuronal Activity},
  author    = {Carzaniga, Francesco S. and Hersche, Michael and Sebastian, Abu and
               Schindler, Kaspar and Rahimi, Abbas},
  booktitle = {The Fourteenth International Conference on Learning Representations},
  year      = {2026},
  url       = {https://openreview.net/forum?id=5M1YOW3bRq}
}

@article{zada2025podcast,
  title   = {The {``Podcast''} {ECoG} Dataset for Modeling Neural Activity
             During Natural Language Comprehension},
  author  = {Zada, Zaid and Nastase, Samuel A. and Aubrey, Bobbi and Jalon, Itamar and
             Michelmann, Sebastian and Wang, Haocheng and Hasenfratz, Liat and
             Doyle, Werner and Friedman, Daniel and Dugan, Patricia and Melloni, Lucia and
             Devore, Sasha and Flinker, Adeen and Devinsky, Orrin and Goldstein, Ariel and
             Hasson, Uri},
  journal = {Scientific Data},
  volume  = {12},
  number  = {1},
  pages   = {1135},
  year    = {2025},
  doi     = {10.1038/s41597-025-05462-2}
}

@article{hatano2026validation,
  title   = {Internal and External Validation of Comprehensive High-Frequency
             Activity Biomarkers for Epilepsy Surgery},
  author  = {Hatano, Keisuke and Kuroda, Naoto and Uda, Hiroshi and Sakakura, Kazuki and
             Cools, Michael J. and Luat, Aimee F. and Osawa, Shin-Ichiro and Nemoto, Hitoshi and
             Ukishiro, Kazushi and Endo, Hidenori and Nakasato, Nobukazu and Takayama, Yutaro and
             Iijima, Keiya and Iwasaki, Masaki and Asano, Eishi},
  journal = {Clinical Neurophysiology},
  volume  = {190},
  pages   = {2111981},
  year    = {2026},
  doi     = {10.1016/j.clinph.2026.2111981}
}

@article{jayakar2016diagnostic,
  title={Diagnostic utility of invasive EEG for epilepsy surgery: indications, modalities, and techniques},
  author={Jayakar, Prasanna and Gotman, Jean and Harvey, A Simon and Palmini, Andr{\'e} and Tassi, Laura and Schomer, Donald and Dubeau, Francois and Bartolomei, Fabrice and Yu, Alice and Kr{\v{s}}ek, Pavel and others},
  journal={Epilepsia},
  volume={57},
  number={11},
  pages={1735--1747},
  year={2016},
  publisher={Wiley Online Library}
}

@article{bernabei2023quantitative,
  title={Quantitative approaches to guide epilepsy surgery from intracranial {EEG}},
  author={Bernabei, John M and Li, Adam and Revell, Andrew Y and Smith, Rachel J and Gunnarsdottir, Kristin M and Ong, Ian Z and Davis, Kathryn A and Sinha, Nishant and Sarma, Sridevi and Litt, Brian},
  journal={Brain},
  volume={146},
  number={6},
  pages={2248--2258},
  year={2023},
  doi={10.1093/brain/awad007},
  publisher={Oxford University Press US}
}

@article{makin2020translation,
  author  = {Makin, Joseph G. and Moses, David A. and Chang, Edward F.},
  title   = {Machine Translation of Cortical Activity to Text with an
             Encoder--Decoder Framework},
  journal = {Nature Neuroscience},
  volume  = {23},
  number  = {4},
  pages   = {575--582},
  year    = {2020},
  doi     = {10.1038/s41593-020-0608-8}
}

@article{rosenow2001presurgical,
  title={Presurgical evaluation of epilepsy},
  author={Rosenow, Felix and L{\"u}ders, Hans},
  journal={Brain},
  volume={124},
  number={9},
  pages={1683--1700},
  year={2001},
  publisher={Oxford University Press},
  doi={10.1093/brain/124.9.1683}
}

@article{kane2017glossary,
  title={A revised glossary of terms most commonly used by clinical electroencephalographers and updated proposal for the report format of the EEG findings. Revision 2017},
  author={Kane, Nick and Acharya, Jayant and Beniczky, Sandor and Caboclo, Luis and Finnigan, Simon and Kaplan, Peter W and Shibasaki, Hiroshi and Pressler, Ronit and Van Putten, Michel JAM},
  journal={Clinical neurophysiology practice},
  volume={2},
  pages={170--185},
  year={2017},
  publisher={Elsevier},
  doi       = {10.1016/j.cnp.2017.07.002}
}

@article{zijlmans2012high,
  title   = {High-frequency oscillations as a new biomarker in epilepsy},
  author  = {Zijlmans, Maeike and Jiruska, Premysl and Zelmann, Rina and Leijten, Frans S. S. and Jefferys, John G. R. and Gotman, Jean},
  journal = {Annals of Neurology},
  volume  = {71},
  number  = {2},
  pages   = {169--178},
  year    = {2012},
  doi     = {10.1002/ana.22548}
}

@article{daida2026developmental,
  title   = {Developmental profile of physiological high-frequency oscillations in the human brain},
  author  = {Daida, Atsuro and Kanai, Sotaro and Zhang, Yipeng and Duan, Chenda and Kuroda, Naoto and Monsoor, Tonmoy and Ding, Yuanyi and Kaneko, Kikuko and Qiao, Joe and Hussain, Shaun A. and Fallah, Aria and Salamon, Noriko and Sankar, Raman and Staba, Richard J. and Engel, Jerome and Speier, William and Asano, Eishi and Roychowdhury, Vwani and Nariai, Hiroki},
  journal = {NeuroImage},
  volume  = {336},
  pages   = {122017},
  year    = {2026},
  doi     = {10.1016/j.neuroimage.2026.122017}
}

@article{martinet2017slow,
  title={Human seizures couple across spatial scales through travelling wave dynamics},
  author={Martinet, Louis-Emmanuel and Fiddyment, Grant and Madsen, Joseph R and Eskandar, Emad N and Truccolo, Wilson and Eden, Uri T and Cash, Sydney S and Kramer, Mark A},
  journal={Nature communications},
  volume={8},
  number={1},
  pages={14896},
  year={2017},
  publisher={Nature Publishing Group UK London},
  doi       = {10.1038/ncomms14896}
}

@inproceedings{laionclap2023,
  title={Large-scale contrastive language-audio pretraining with feature fusion and keyword-to-caption augmentation},
  author={Wu, Yusong and Chen, Ke and Zhang, Tianyu and Hui, Yuchen and Berg-Kirkpatrick, Taylor and Dubnov, Shlomo},
  booktitle={ICASSP 2023-2023 IEEE International Conference on Acoustics, Speech and Signal Processing (ICASSP)},
  pages={1--5},
  year={2023},
  organization={IEEE}
}

@article{taylor2022normative,
  author  = {Taylor, Peter N and Papasavvas, Christoforos A and Owen, Thomas W and Schroeder, Gabrielle M and Hutchings, Frances E and Chowdhury, Fahmida A and Diehl, Beate and Duncan, John S and McEvoy, Andrew W and Miserocchi, Anna and de Tisi, Jane and Vos, Sjoerd B and Walker, Matthew C and Wang, Yujiang},
  title   = {Normative brain mapping of interictal intracranial {EEG} to localize epileptogenic tissue},
  journal = {Brain},
  volume  = {145},
  number  = {3},
  pages   = {939--949},
  year    = {2022},
  doi     = {10.1093/brain/awab380}
}

@article{lachaux2012high,
  author  = {Lachaux, Jean-Philippe and Axmacher, Nikolai and Mormann, Florian and Halgren, Eric and Crone, Nathan E.},
  title   = {High-frequency neural activity and human cognition: Past, present and possible future of intracranial {EEG} research},
  journal = {Progress in Neurobiology},
  volume  = {98},
  number  = {3},
  pages   = {279--301},
  year    = {2012},
  doi     = {10.1016/j.pneurobio.2012.06.008}
}

@article{hamilton2018spatial,
  author  = {Hamilton, Liberty S. and Edwards, Erik and Chang, Edward F.},
  title   = {A Spatial Map of Onset and Sustained Responses to Speech in the Human Superior Temporal Gyrus},
  journal = {Current Biology},
  volume  = {28},
  number  = {12},
  pages   = {1860--1871.e4},
  year    = {2018},
  doi     = {10.1016/j.cub.2018.04.033}
}

@article{tang2017intonational,
  title={Intonational speech prosody encoding in the human auditory cortex},
  author={Tang, Claire and Hamilton, Liberty S and Chang, Edward F},
  journal={Science},
  volume={357},
  number={6353},
  pages={797--801},
  year={2017},
  publisher={American Association for the Advancement of Science},
  doi       = {10.1126/science.aam8577}
}

@article{potes2012dynamics,
  author  = {Potes, Cristhian and Gunduz, Aysegul and Brunner, Peter and Schalk, Gerwin},
  title   = {Dynamics of electrocorticographic ({ECoG}) activity in human temporal and frontal cortical areas during music listening},
  journal = {NeuroImage},
  volume  = {61},
  number  = {4},
  pages   = {841--848},
  year    = {2012},
  doi     = {10.1016/j.neuroimage.2012.04.022}
}

@article{nejedly2020multicenter,
  author  = {Nejedly, Petr and Kremen, Vaclav and Sladky, Vladimir and Cimbalnik, Jan and Klimes, Petr and Plesinger, Filip and Mivalt, Filip and Travnicek, Vojtech and Viscor, Ivo and Pail, Martin and Halamek, Josef and Brinkmann, Benjamin H. and Brazdil, Milan and Jurak, Pavel and Worrell, Gregory},
  title   = {Multicenter intracranial {EEG} dataset for classification of graphoelements and artifactual signals},
  journal = {Scientific Data},
  volume  = {7},
  number  = {1},
  pages   = {179},
  year    = {2020},
  doi     = {10.1038/s41597-020-0532-5}
}

@article{lawhern2018eegnet,
  author  = {Lawhern, Vernon J and Solon, Amelia J and Waytowich, Nicholas R and Gordon, Stephen M and Hung, Chou P and Lance, Brent J},
  title   = {{EEGNet}: a compact convolutional neural network for {EEG}-based brain--computer interfaces},
  journal = {Journal of Neural Engineering},
  volume  = {15},
  number  = {5},
  pages   = {056013},
  year    = {2018},
  doi     = {10.1088/1741-2552/aace8c}
}

@article{huang2023emotion,
  author  = {Huang, Zhentao and Ma, Yahong and Wang, Rongrong and Li, Weisu and Dai, Yongsheng},
  title   = {A Model for {EEG}-Based Emotion Recognition: {CNN-Bi-LSTM} with Attention Mechanism},
  journal = {Electronics},
  volume  = {12},
  number  = {14},
  pages   = {3188},
  year    = {2023},
  doi     = {10.3390/electronics12143188}
}

@inproceedings{nie2023patchtst,
  author    = {Nie, Yuqi and Nguyen, Nam H. and Sinthong, Phanwadee and Kalagnanam, Jayant},
  title     = {A Time Series is Worth 64 Words: Long-term Forecasting with Transformers},
  booktitle = {The Eleventh International Conference on Learning Representations},
  year      = {2023}
}

@inproceedings{wu2023timesnet,
  author    = {Wu, Haixu and Hu, Tengge and Liu, Yong and Zhou, Hang and Wang, Jianmin and Long, Mingsheng},
  title     = {{TimesNet}: Temporal {2D}-Variation Modeling for General Time Series Analysis},
  booktitle = {The Eleventh International Conference on Learning Representations},
  year      = {2023}
}

@article{ding2025pyhfo,
  author  = {Ding, Yuanyi and Zhang, Yipeng and Duan, Chenda and Daida, Atsuro and Zhang, Yun and Kanai, Sotaro and Lu, Mingjian and Hussain, Shaun and Staba, Richard J and Nariai, Hiroki and Roychowdhury, Vwani},
  title   = {{PyHFO} 2.0: an open-source platform for deep learning--based clinical high-frequency oscillations analysis},
  journal = {Journal of Neural Engineering},
  volume  = {22},
  number  = {5},
  pages   = {056040},
  year    = {2025},
  doi     = {10.1088/1741-2552/ae10e0}
}

@article{zhang2022refining,
  author  = {Zhang, Yipeng and Lu, Qiujing and Monsoor, Tonmoy and Hussain, Shaun A. and Qiao, Joe X. and Salamon, Noriko and Fallah, Aria and Sim, Myung Shin and Asano, Eishi and Sankar, Raman and Staba, Richard J. and Engel, Jerome and Speier, William and Roychowdhury, Vwani and Nariai, Hiroki},
  title   = {Refining epileptogenic high-frequency oscillations using deep learning: a reverse engineering approach},
  journal = {Brain Communications},
  volume  = {4},
  number  = {1},
  pages   = {fcab267},
  year    = {2022},
  doi     = {10.1093/braincomms/fcab267}
}

@article{lachaux1999measuring,
  title   = {Measuring phase synchrony in brain signals},
  author  = {Lachaux, Jean-Philippe and Rodriguez, Eugenio and Martinerie, Jacques and Varela, Francisco J.},
  journal = {Human Brain Mapping},
  volume  = {8},
  number  = {4},
  pages   = {194--208},
  year    = {1999}
}

@article{bruns2000amplitude,
  title   = {Amplitude envelope correlation detects coupling among incoherent brain signals},
  author  = {Bruns, Andreas and Eckhorn, Reinhard and Jokeit, Hennric and Ebner, Alois},
  journal = {NeuroReport},
  volume  = {11},
  number  = {7},
  pages   = {1509--1514},
  year    = {2000},
  doi     = {10.1097/00001756-200005150-00029}
}

@inproceedings{esteller2001line,
  title     = {Line length: an efficient feature for seizure onset detection},
  author    = {Esteller, Rosana and Echauz, Javier and Tcheng, Tung and Litt, Brian and Pless, Benjamin},
  booktitle = {Proceedings of the 23rd Annual International Conference of the IEEE Engineering in Medicine and Biology Society},
  volume    = {2},
  pages     = {1707--1710},
  year      = {2001}
}
\bibliographystyle{plainnat}

\clearpage
\appendix
\raggedbottom
\renewcommand{\topfraction}{0.95}
\renewcommand{\bottomfraction}{0.95}
\renewcommand{\textfraction}{0.05}
\renewcommand{\floatpagefraction}{0.85}
\setcounter{topnumber}{5}
\setcounter{bottomnumber}{5}
\setcounter{totalnumber}{10}

\makeatletter
\newenvironment{appendixtable}{%
  \par\addvspace{10pt}\noindent\begin{minipage}{\textwidth}%
  \def\@captype{table}%
}{%
  \end{minipage}\par\addvspace{10pt}%
}
\makeatother

\begin{center}
\Large\textbf{Appendix:}
\end{center}
\vspace{0.5em}

\etocdepthtag.toc{mtappendix}
\begingroup
\hypersetup{hidelinks,linktoc=all}
\etocsettagdepth{mtchapter}{none}
\etocsettagdepth{mtappendix}{subsubsection}
\etocsetnexttocdepth{subsubsection}
\etocsettocstyle{\section*{Contents of Appendix}}{}
\tableofcontents
\endgroup

\clearpage

\section{Data and Signal Preparation}
\label{app:implementation}
\label{app:data-signal-preparation}

\subsection{Pretraining Data and Partitions}
\label{sec:data}

PHASE-T uses 568 recordings from Omni-iEEG~\citep{duan2026omniieeg},
the Hatano Detroit release~\citep{hatano2026validation}, NYU Podcast
ECoG~\citep{zada2025podcast}, and BrainTreebank~\citep{wang2024brain}
(Table~\ref{tab:source-composition}). Sampling is balanced across
participants within each source, not by duration.

Pretraining uses the official Omni-iEEG training partition, the Hatano Detroit non-REM recordings, the NYU Podcast ECoG recordings, and BrainTreebank's pretraining split.
No downstream evaluation data enter pretraining.
Omni-iEEG test participants are excluded from both clinical sources. Pretraining also contains no ictal recordings: all ictal recordings in Omni-iEEG, including every seizure analyzed in Section~\ref{sec:readings}, are excluded, and the remaining sources contribute interictal clinical monitoring or naturalistic listening sessions.
The Tohoku and NCNP cohorts of the Hatano release and the MAYO and FNUSA cohorts are used only for evaluation.
For BrainTreebank, we follow BaRISTA's recording split~\citep{oganesian2025barista}: 17 recordings for pretraining, 2 for validation, and 7 for testing.

\begin{appendixtable}
\centering
\caption{\textbf{Pretraining sources.}
Recs, recordings. Sampling shares are proportional to each source's participant count, and
channel-hours are counted before artifact rejection.
Omni-iEEG and Hatano share 76 participants, so the corpus includes 222 distinct participants.}
\label{tab:source-composition}
\label{tab:source-scale-governance}
\small
\setlength{\tabcolsep}{3.5pt}
\renewcommand{\arraystretch}{1.10}
\begin{tabularx}{\textwidth}{@{}lrrrrXX@{}}
\toprule
Source & Participants & Recs & Channel-hours & Sampling & Recording context & Evaluation use \\
\midrule
Omni-iEEG & 151 & 399 & 5,235 & 50.67\% &
Clinical monitoring; official training partition &
Test participants and tasks \\
Hatano & 129 & 144 & 10,315 & 43.29\% &
Clinical monitoring; Detroit non-REM sleep &
Tohoku and NCNP cohorts \\
NYU Podcast & 8 & 8 & 632 & 2.68\% &
Naturalistic listening &
Pretraining only \\
BrainTreebank & 10 & 17 & 5,434 & 3.36\% &
Naturalistic listening; BaRISTA pretraining split &
7 test recordings \\
\bottomrule
\end{tabularx}
\end{appendixtable}

\subsection{Data Preparation and Signal Preprocessing}
\label{app:signal-preprocessing}

Recordings are converted into single-channel waveforms in microvolts on a common 1\,kHz grid, retaining their native clinical referencing, acquisition bandwidth, and per-sample validity masks.

The deterministic filtering in Table~\ref{tab:signal-contract} is applied when
estimating target statistics, during pretraining, and at inference; the
table's bandwidth augmentation is used only during pretraining.

\begin{appendixtable}
\centering
\caption{\textbf{Shared preprocessing for training and inference.}}
\label{tab:signal-contract}
\small
\renewcommand{\arraystretch}{1.12}
\begin{tabularx}{\textwidth}{@{}p{0.26\textwidth}X@{}}
\toprule
Parameter & Setting \\
\midrule
Waveform units and grid &
Microvolts at 1 kHz; released reference convention retained \\
Pretraining crop &
60 seconds (60,000 samples) \\
High-pass filter &
Second-order Butterworth, 2-Hz corner \\
Mains-notch filters &
50 and 60 Hz, each with quality factor $Q=30$ \\
Deterministic filtering &
Zero-phase product of the squared-magnitude filter responses, applied in
the frequency domain after reflection padding \\
Input amplitude transform &
$u_t=\operatorname{asinh}(x_t/s_0)$ for filtered waveforms with scale parameter
$s_0=20\,\mu\mathrm V$, preserving absolute signal amplitudes \\
Bandwidth augmentation &
Probability 0.25 for crops with usable acquisition bandwidth of at least 240 Hz \\
Augmentation low-pass &
Cutoff sampled uniformly from 200--220 Hz, followed by a 20-Hz raised-cosine
transition; reflection padding of 1 second on each side \\
Augmented views and targets &
Teacher, student, and physiological targets use the same bandwidth-reduced
waveform; the sampled cutoff becomes its effective target bandwidth \\
Frequency eligibility &
Wavelet center frequency or Fourier-band upper edge must not exceed the
effective acquisition bandwidth \\
\bottomrule
\end{tabularx}
\end{appendixtable}

\section{PHASE-T Architecture and Pretraining}
\label{app:model-pretraining-details}

\subsection{Temporal Architecture}

For 60-second, 1\,kHz crops, a convolutional tokenizer embeds overlapping
128-sample patches at a 32-sample stride, with 48 zero-padded samples on
each side. GELU, linear projection, and RMS normalization produce
$N=1{,}875$ tokens of width 384, spaced 32\,ms apart. Eight fine-scale
Transformer blocks use six heads and 256-token local attention windows,
shifted by 128 tokens in alternating blocks, yielding
$H_{\mathrm{fine}}\in\mathbb R^{1{,}875\times384}$.

Convolutional pooling with kernel eight and stride four maps the fine
sequence to 469 tokens of width 768. For the 1,875-token input, pooling
reflect-pads two tokens on the left and three on the right. Ten coarse
blocks with 12 heads use full bidirectional attention, yielding
$H_{\mathrm{coarse}}\in\mathbb R^{469\times768}$. A learned transposed
convolution with the same kernel and stride returns these tokens to the
fine grid; the asymmetric padding is cropped away. The result is added
to a 384-to-768 projection of $H_{\mathrm{fine}}$ and refined by two
12-head blocks with 256-token windows, the second shifted by 128 tokens.
This is the $Z_c$ of Equation~\ref{eq:refinement-decoder}, with
$N=1{,}875$ and $d=768$.

All Transformer blocks use pre-normalized residual connections,
RMS normalization, rotary position embeddings, and SwiGLU
feed-forward layers.
The SwiGLU hidden dimension is $8/3$ times the block dimension:
1,024 in the fine-scale encoder and 2,048 in the coarse-scale
encoder and refinement decoder.
The encoder comprises 106.7M parameters. The latent predictor and
physiological decoder are two-layer GELU networks on $Z_c$.

\subsection{Physiological Target Construction}
\label{app:physiological-targets}

At each token position $t$, we construct an 84-dimensional physiological
target comprising wavelet power, amplitude and local variation,
and short-window high-frequency power:
\[
q_t=[w_t;a_t;h_t]\in\mathbb R^{84},
\qquad
w_t\in\mathbb R^{64},\quad
a_t\in\mathbb R^{4},\quad
h_t\in\mathbb R^{16}.
\]
All targets are computed from unmasked waveforms in microvolts after
signal preprocessing and any pretraining bandwidth augmentation,
but before the asinh input amplitude transform in Table~\ref{tab:signal-contract}.

\paragraph{Wavelet power.}
The spectral target $w_t$ describes power at 64 center frequencies:
24 logarithmically spaced frequencies from 2 to 80\,Hz and
40 linearly spaced frequencies above 80 up to 450\,Hz.
We use complex Morlet wavelets, each with a Gaussian frequency response
and temporal standard deviation $\sigma(f)=3/f$.
To obtain an analytic response with amplitude-preserving
normalization, the frequency-domain filter is set to zero at
negative frequencies and multiplied by two at positive frequencies.
Reflection padding extends the waveform by 6 seconds on each side,
corresponding to four temporal standard deviations of the
lowest-frequency wavelet.

For each frequency, the squared magnitude of the wavelet coefficients
is averaged within consecutive 32-sample windows to align the power
measurements with the token grid.
The resulting power $P$ is log-transformed as
\[
\widetilde P=\log\!\left(1+\frac{P}{P_0}\right),
\qquad P_0=(1\,\mu\mathrm V)^2.
\]

\paragraph{Amplitude and local variation.}
The target $a_t$ contains four waveform descriptors:
centered root-mean-square amplitude, mean absolute deviation,
peak-to-peak amplitude, and mean absolute successive difference~\citep[line length;][]{esteller2001line}.
These descriptors are computed over 128-sample patches using the
same stride and zero padding as the tokenizer.
Each descriptor $a$ is log-transformed as
\[
\widetilde a=\log\!\left(1+\frac{a}{s_0}\right),
\qquad s_0=20\,\mu\mathrm V.
\]

\paragraph{Local high-frequency power.}
The target $h_t$ captures spectral content using the
same 128-sample patches.
Each patch is multiplied by a symmetric Hann window before
Fourier transformation.
Squared Fourier magnitudes are normalized by the window energy,
and positive-frequency bins other than the Nyquist bin are
doubled to obtain a one-sided power spectrum; the DC bin is unchanged.
The spectrum is summarized by averaging power across bins within
each of 16 bands: eight equal-width bands over 80--250\,Hz and
eight over 250--450\,Hz, in the ripple and fast-ripple ranges~\citep{zijlmans2012high}.
Each band-power measurement $P$ is log-transformed as
$\log(1+P/s_0^2)$.

\paragraph{Target standardization.}
Each log-transformed target feature is standardized using a fixed
mean and standard deviation estimated from the participant-balanced
pretraining mixture.
These statistics are computed using the same deterministic
signal-filtering pipeline used by the encoder and remain fixed
throughout pretraining.

\subsection{Masking and Temporal Objectives}
\label{app:temporal-objectives}

Temporal pretraining combines latent prediction at masked positions
with physiological supervision at positions selected according to
target type and waveform visibility.

\paragraph{Temporal masking.}
For each microbatch, we sample temporal spans covering 55\% of token
positions, denoted by $\mathcal M$.
Each span length is sampled uniformly from one of three integer
ranges, with range-selection probabilities specified in
Table~\ref{tab:temporal-pretraining-config}.
Spans may overlap, and the final span may be shortened to reach
the specified number of prediction positions.

To prevent overlapping patches from exposing masked waveform samples,
we first zero all samples covered by the tokenizer patches at
positions in $\mathcal M$.
Every token whose patch overlaps these zeroed samples is replaced
with a learned mask embedding.
We denote the fully visible positions by $\mathcal V$.
The positions outside $\mathcal M\cup\mathcal V$ correspond
to partially masked patches adjacent to the prediction spans.

\paragraph{Latent prediction.}
The masked student and unmasked teacher produce representations
$Z^{\mathrm S}$ and $Z^{\mathrm T}$, respectively.
Both views use the same deterministic preprocessing and the same
bandwidth augmentation as the physiological targets.
The teacher is evaluated in inference mode, and its parameters are
updated by an exponential moving average of the student.

At each position in $\mathcal M$, the latent predictor $p_\phi$
maps the student representation to the corresponding teacher target.
The prediction loss is
\begin{equation}
\label{eq:temporal-jepa}
\mathcal L_{\mathrm{latent}}^{\mathrm T}
=\frac{1}{d|\mathcal M|}\sum_{t\in\mathcal M}
\left\|
\operatorname{LN}(p_\phi(Z^{\mathrm S}_t))
-\operatorname{sg}[\operatorname{LN}(Z^{\mathrm T}_t)]
\right\|_2^2,
\end{equation}
where $d$ is the embedding dimension,
$\operatorname{LN}$ denotes feature normalization without learned
affine parameters, and $\operatorname{sg}$ denotes stop-gradient.
The teacher update schedule is specified below.

\paragraph{Physiological supervision and target visibility.}
The physiological decoder $g_\psi$ maps the same student representations
to predictions of $q_t=[w_t;a_t;h_t]$.
Wavelet-power and amplitude losses are evaluated at both masked
prediction positions and fully visible positions,
$t\in\mathcal M\cup\mathcal V$.
The local Fourier-power loss is evaluated only at fully visible
positions, $t\in\mathcal V$.

Partially masked patches outside $\mathcal M$ are excluded from
the physiological loss.

\paragraph{Bandwidth-dependent target selection.}
Frequency-dependent targets contribute to the loss only when supported
by the recording's effective bandwidth after any bandwidth augmentation.
A wavelet-power target is included when its center frequency lies
within this bandwidth; a Fourier-power band is included when its
upper edge lies within it.
This selection prevents supervision at frequencies unavailable in
the original recording or removed by augmentation.

\paragraph{Physiological loss and joint optimization.}
For each target group $b\in\{w,a,h\}$, let $\mathcal I_b$ contain
the token--feature pairs satisfying the visibility and bandwidth
conditions above, and let $\mathcal B$ denote the groups with at
least one valid pair.
We first average the loss within each group and then average across
groups, giving each active target group equal weight:
\begin{equation}
\label{eq:physiological-loss}
\mathcal L_{\mathrm{phys}}
=\frac{1}{|\mathcal B|}\sum_{b\in\mathcal B}
\frac{1}{|\mathcal I_b|}\sum_{(t,k)\in\mathcal I_b}
\rho\!\left(g_\psi(Z^{\mathrm S}_t)_{b,k}-q_{t,b,k}\right),
\end{equation}
where $\rho$ is the Smooth L1 loss with transition parameter one.
The student encoder, latent predictor, and physiological decoder are
optimized jointly using
$\mathcal L_{\mathrm{latent}}^{\mathrm T}+\lambda_{\mathrm{phys}}\mathcal L_{\mathrm{phys}}$,
as defined in Equation~\ref{eq:temporal-objective}.

\paragraph{Ablation variants.}
We compare PHASE-T with three variants in Table~\ref{tab:mechanism-ablations},
each pretrained with the settings in Table~\ref{tab:temporal-pretraining-config}
apart from the changes described below.
Latent-only sets $\lambda_{\mathrm{phys}}=0$.
Latent+Recon replaces physiological supervision with
Brant-style reconstruction~\citep{zhang2023brant} of clean 128-sample waveform patches aligned to the token
grid. Targets use the preprocessed, bandwidth-augmented waveform before
masking and the input amplitude transform, standardized by a fixed scalar mean
and standard deviation estimated from the training samples.
A two-layer GELU decoder predicts each patch from its student token.
As in masked autoencoders~\citep{he2022mae}, mean squared error is computed only at the
requested masked positions $\mathcal M$, averaged over in-crop samples, and added to
$\mathcal L_{\mathrm{latent}}^{\mathrm T}$ with unit weights on both losses.
Both physiological and reconstruction targets are standardized using fixed
training-set statistics, and both auxiliary losses enter the latent-prediction
objective with weight one. The comparison shares the encoder architecture,
pretraining settings, and downstream probing protocol, while each auxiliary
objective follows its specified target definition, supervision locations,
and loss function.
No-skip removes $\operatorname{Proj}(H_{\mathrm{fine}})$
from Equation~\ref{eq:refinement-decoder}, leaving the refinement decoder
to receive only upsampled coarse features.

\subsection{Pretraining Settings}
\label{app:temporal-training}

We optimize the student encoder and prediction heads with AdamW,
applying weight decay to matrix weights while excluding biases,
normalization gains, and the learned mask embedding.

After each successful optimizer update $s$, the teacher parameters
$\bar\theta$ are updated from the student parameters $\theta$:
\[
\bar\theta \leftarrow m(s)\bar\theta+[1-m(s)]\theta.
\]
The EMA momentum increases from $m_{\mathrm{base}}$ to
$m_{\mathrm{final}}$ on a cosine schedule.

All downstream evaluations use the EMA teacher checkpoint at
60,000 updates, corresponding to 11.52 million sampled training
crops, as either a frozen encoder or the initialization for
task-specific fine-tuning.
Table~\ref{tab:temporal-pretraining-config} summarizes the
optimization and masking settings.

\begin{appendixtable}
\centering
\caption{\textbf{Temporal pretraining with multiscale masking and a moving-average teacher.}
Mask-span lengths are in fine-grid tokens; schedules are in optimizer updates.}
\label{tab:temporal-pretraining-config}
\small
\renewcommand{\arraystretch}{1.12}
\begin{tabularx}{\textwidth}{@{}p{0.34\textwidth}X@{}}
\toprule
Parameter & Setting \\
\midrule
Requested mask fraction & 55\% of fine-grid tokens \\
Short mask spans & 2--4 tokens, probability 0.35 \\
Medium mask spans & 16--62 tokens, probability 0.45 \\
Long mask spans & 156--312 tokens, probability 0.20 \\
Optimizer & AdamW; $(\beta_1,\beta_2)=(0.9,0.95)$, $\epsilon=10^{-10}$ \\
Weight decay & 0.1 on matrix weights \\
Learning rate & Peak $3\times10^{-4}$ \\
Learning-rate schedule & 1,000 linear warm-up updates, then cosine decay to $7.7\times10^{-5}$ at 60,000 updates \\
Global batch & 192 crops (48 crops per GPU across 4 GPUs) \\
Training hardware & 4$\times$ NVIDIA PRO 6000; 2.8 GPU-days for 60K updates \\
Numerical precision & Bfloat16 autocast \\
Gradient clipping & Global gradient norm of 1.0 \\
EMA momentum & Cosine from $m_{\mathrm{base}}=0.996$ to $m_{\mathrm{final}}=0.9989$ over 60,000 updates \\
Temporal objective weights & Latent: 1; physiological: $\lambda_{\mathrm{phys}}=1$ \\
\bottomrule
\end{tabularx}
\end{appendixtable}

\section{PHASE-ST Architecture and Pretraining}
\label{app:channel-set-method-details}

\subsection{Spatial Residual Architecture}

\paragraph{Frozen temporal interface.}
For a synchronized set of $C$ channels, the frozen PHASE-T teacher encodes
each waveform independently. Stacking its dense outputs gives
$X\in\mathbb R^{C\times N\times768}$, with $N=1{,}875$ for a 60-second
input and $N=156$ for a five-second input at 1\,kHz. Both training and
inference use PHASE-T's waveform preprocessing. Within
the spatial branch, we subtract a fixed training-mean template
$\mu^{(L)}\in\mathbb R^{N\times768}$ matched to the input duration $L$.
This position-dependent template is shared across channels and retained
with the model; one template per input duration is estimated once with
frozen PHASE-T from training windows (two per participant, up to eight
channels each), weighted as in pretraining sampling. The residual branch is centered without feature rescaling;
the original tokens $X$ remain the skip input and clean prediction target.

A bias-free projection $W_\downarrow$ maps centered tokens from width 768
to 256. Four Transformer blocks then apply self-attention across the $C$
channels independently at each time position. The same block parameters
are used at all positions. Each block has pre-attention RMS normalization,
four attention heads of width 64, and a pre-normalized SwiGLU feed-forward
layer with hidden dimension 512. Both sublayers have residual connections;
dropout is zero. Writing $F_\ell$ for spatial block $\ell$, the complete
update is
\begin{equation}
\label{eq:spatial-residual}
\begin{aligned}
U^{(0)}_{:,t}&=W_\downarrow(X_{:,t}-\mu^{(L)}_t),\\
U^{(\ell)}_{:,t}&=F_\ell(U^{(\ell-1)}_{:,t}),\qquad \ell=1,\ldots,4,\\
H_{:,t}&=X_{:,t}+W_\uparrow\bigl(U^{(4)}_{:,t}-U^{(0)}_{:,t}\bigr),
\end{aligned}
\end{equation}
where the bias-free $W_\uparrow$ maps width 256 back to 768. Attention
uses the current states of all channels, including the query channel,
as keys and values at every layer. The module adds no temporal attention,
positional rotation, channel-index embedding, or electrode-coordinate
input; temporal context is already carried by PHASE-T. Consequently,
permuting the channels permutes $H$ in the same way, and the number of
channels can vary without changing the parameters.

Attention-output and feed-forward-output weights start at zero, giving
$U^{(4)}=U^{(0)}$ and $H=X$ at initialization. The spatial module contains
3,016,704 trainable parameters, including its two projections. The latent
predictor and relation heads bring the pretraining total to 4,260,560;
the frozen temporal encoder, frozen physiological decoder, and EMA copy
are excluded from this count.

\subsection{Latent Prediction and Physiological Supervision}

\paragraph{Latent prediction of fixed temporal targets.}
Independent multiscale masks are sampled for each channel, using the
55\% requested fraction and span distribution in
Table~\ref{tab:temporal-pretraining-config}. Let $\mathcal M$ contain the
requested channel--time positions. After PHASE-T's waveform preprocessing,
we zero waveform samples covered by those tokens' receptive fields and
mark every overlapping tokenizer patch as masked. Frozen PHASE-T encodes
both the clean and masked waveforms. Its masked outputs
$\widetilde X$ retain the contextual states at all positions, including
the masked positions, and feed the spatial module to produce
$\widetilde H$. The retained within-channel context distinguishes channels
while spatial attention supplies synchronized cross-channel information.

A two-layer predictor $p$ (768--768--768, with GELU between the linear
layers) maps $\widetilde H$ to clean PHASE-T targets. With non-affine,
per-token feature normalization $\operatorname{LN}$, the loss is
\begin{equation}
\label{eq:spatial-context}
\mathcal L_{\mathrm{latent}}^{\mathrm{ST}}
=\frac{1}{768|\mathcal M|}\sum_{(c,t)\in\mathcal M}
\left\|\operatorname{LN}(p(\widetilde H_{c,t}))
-\operatorname{sg}[\operatorname{LN}(X_{c,t})]\right\|_2^2.
\end{equation}
The target is fixed by the frozen temporal encoder. Only the requested
mask positions are scored; the additional overlap halo is excluded.

\paragraph{Frozen physiological supervision.}
The 84-output physiological decoder saved with PHASE-T is applied directly
to $\widetilde H$. Its weights and target statistics remain fixed, while
gradients through the decoder train the spatial module.
Targets are computed from the clean, preprocessed waveforms using the
definitions, transforms, and fixed standardization in
Appendix~\ref{app:physiological-targets}; supervision follows the visibility,
bandwidth, and loss rules in Appendix~\ref{app:temporal-objectives}.

\subsection{Pairwise Relation Targets and Loss}

\paragraph{Pairwise relation targets.}
Targets are measured from the continuous clean waveforms after PHASE-T's
waveform preprocessing and temporal mean removal. Fourth-order
Butterworth band-pass filters applied forward and backward, followed by
the Hilbert transform, define analytic signals
\begin{equation}
\label{eq:analytic-polar}
z_{c,b}(t)=x_{c,b}(t)+i\,\mathcal H\{x_{c,b}(t)\}
=A_{c,b}(t)e^{i\phi_{c,b}(t)}.
\end{equation}
For bands 1--4, 4--8, 8--13, 13--30, 30--80, 80--250, and 250--500\,Hz,
we measure the phase-locking value
$|\langle e^{i(\phi_{i,b}-\phi_{j,b})}\rangle_t|$ and the Pearson
correlation of the amplitude envelopes. The filter's upper edge is capped
at 99\% of Nyquist. Three broadband targets give the signed correlation
at the largest absolute cross-correlation peak within $\pm100$\,ms,
zero-lag correlation, and the peak's signed lag. A band contributes only
when both channels' acquisition bandwidths reach its nominal upper edge.

Phase-locking values are clipped to $[0.001,0.999]$ and logit transformed;
correlations are clipped to $[-0.999,0.999]$ and Fisher transformed.
Lag is divided by 50\,ms. Each coordinate is then standardized using fixed
training moments.

\paragraph{Pairwise relation heads and loss.}
Relation prediction uses the same masked spatial output
$\widetilde H$. For each channel, an affine projection maps its temporal
mean to $r_c\in\mathbb R^{64}$. A symmetric head takes
$[r_i+r_j;\,r_i\odot r_j]$ through a 128--64--16 GELU network.
A separate bias-free 64--64--1 network predicts lag from the difference:
\begin{equation}
\label{eq:pair-heads}
\widehat y^{\mathrm{sym}}_{ij}=h_\eta([r_i+r_j;\,r_i\odot r_j]),
\qquad a_{ij}=W_2\tanh(W_1(r_i-r_j)).
\end{equation}
The symmetric outputs are invariant to channel exchange. For lag, the
standardized prediction is $a_{ij}-\mu_\tau/\sigma_\tau$, where
$\mu_\tau,\sigma_\tau$ are the fixed training moments of the scaled lag
target. Undoing this standardization gives $\sigma_\tau a_{ij}$, which
changes sign when channels are exchanged. Thus antisymmetry holds in
the underlying lag units even when its training mean is nonzero.
We sample 64 distinct unordered pairs per window and randomize their
orientation. Smooth L1 errors are averaged over valid pairs separately
for each window and target coordinate, then over nonempty
window--coordinate groups to give $\mathcal L_{\mathrm{relation}}$.

\subsection{Pretraining and Inference Settings}
\label{app:training-inference}

\paragraph{Channel-set pretraining (PHASE-ST).}
Training uses 64-channel, 60-second sets sampled from the four-source
training corpus: each source contributes a fixed quota of sets, and within a
source a participant is drawn uniformly before one of their windows. The global batch
is 128 window sets. AdamW uses learning rate $10^{-4}$,
$(\beta_1,\beta_2)=(0.9,0.95)$, $\epsilon=10^{-8}$, and weight decay
0.05. Learning rate warms up linearly for 1,000 updates and then remains
constant. The spatial module is trained for 5,000 updates, taking approximately
1.9 GPU-days on two NVIDIA PRO 6000 GPUs. Training uses
bfloat16 autocast, float32 residual states, and gradient clipping at norm
1.0. Equation~\ref{eq:spatial-objective} uses weights
$1:1:0.2$ for latent prediction, physiological supervision, and relation losses. All three
losses share the masked spatial output. PHASE-T and its physiological
decoder remain frozen throughout.

An EMA of the spatial module, with momentum rising from 0.996 to 0.9966 along
a cosine ramp over the 5,000 updates, is updated after each
successful optimizer step and supplies the exported spatial representation; the prediction targets
remain the clean frozen PHASE-T tokens. Both reported auditory rows use the
resulting step-5,000 spatial EMA checkpoint together with the same step-60,000
temporal teacher.

\paragraph{Inference and exported representations.}
For PHASE-ST, inference uses unmasked inputs and an EMA of the spatial
module, with pretraining heads removed. PHASE-T supplies the
temporal tokens.
The exported $H$ retains every channel and time position.
Temporal averaging yields one vector per channel. Linear probing and fine-tuning for auditory decoding are
described in Appendix~\ref{braintreebank-auditory-protocol}.

\clearpage
\section{Evaluation Protocols and Extended Results}
\label{app:downstream-implementation}
\label{app:details}

\subsection{Shared Encoder Settings}
\label{reported-encoder-and-readout-protocol}

All PHASE-T evaluations use the same EMA-teacher checkpoint,
saved at step 60,000 of participant-balanced pretraining.
This checkpoint serves as the frozen feature extractor for
task-specific linear probes and the initialization for jointly
fine-tuning the encoder and prediction head.
Both settings retain the preprocessing associated with the
pretrained model.

\subsection{Clinical Prediction across Scales}\label{omni-ieeg-clinical-tasks}

We evaluate clinical prediction following the official Omni-iEEG benchmark formulations, participant-level partitions, and evaluation units~\citep{duan2026omniieeg}, spanning four temporal and clinical prediction scales:
\begin{itemize}[leftmargin=*]
    \item \textbf{Pathological HFO (event level, 570 ms)} classifies 570-ms crops of candidate events, detected with PyHFO~\citep{ding2025pyhfo} and annotated by experts in Omni-iEEG, into artifact, non-spike HFO, or spike-associated HFO, evaluated by macro-F1 across 14,141 held-out test events from 20 participants. Final token representations are averaged across each crop.
    \item \textbf{Ictal period and sleep--awake classification (window level, 60 s)} evaluate 60-second recording windows by averaging channel decision logits within each window. Following the official Omni-iEEG evaluator, macro-F1 is
computed at the threshold maximizing Youden's $J$.
    \item \textbf{Anatomical localization (channel level)} classifies channels into Lobe-5 and Region-12 anatomical targets using standard 60-second segments, evaluated by macro-F1.
    \item \textbf{Pathological channel identification and surgical outcome (channel and participant levels)} follow Omni-iEEG's official pathological-brain-region formulation using 60-second windows and its benchmark-defined evaluators at two complementary clinical levels:
    \begin{itemize}[leftmargin=1.5em]
        \item \emph{Channel-level pathology (ROC-AUC)} assesses discriminative identification of seizure-onset-zone (SOZ) channels across 102 test participants (807 SOZ and 7,297 normal channels), using normal channels from seizure-free participants as controls.
        \item \emph{Participant-level surgical outcome (ROC-AUC)} follows the benchmark's official resection-based evaluation of post-operative seizure freedom (Engel Class~I vs II--IV), which builds on the resection-ratio analyses of prior HFO studies~\citep{zhang2022refining,monsoor2023hfo}.
    \end{itemize}
\end{itemize}

\subsubsection{Linear Probing and Task-Specific Fine-Tuning}
\label{task-specific-fine-tuning}

Following Omni-iEEG's official split, linear probes are fitted on frozen training-set representations with a fixed $\ell_2$ penalty weight $\lambda=0.01$ on the mean cross-entropy, without cross-validation.
Frozen and fine-tuned evaluations, and the ablation probes in Table~\ref{tab:mechanism-ablations}, use the same training rows for each task; the source probe in Table~\ref{tab:external-transfer-detailed} is fitted on the complete official training set.
For task-specific fine-tuning, models are trained for three epochs using AdamW (encoder learning rate $10^{-5}$, head $10^{-4}$, weight decay 0.01, effective batch size 32, bfloat16 precision). Learning rates use a 5\% warm-up, with the first epoch updating only the head.
Pathological-channel training uses class-margin cross-entropy with margin $m=2\log(n_{\mathrm{normal}}/n_{\mathrm{pathological}})$ on the normal class.
Checkpoint selection uses the benchmark validation partitions.

\subsubsection{Released Foundation Checkpoints and External Adapters}

We evaluate three released foundation model checkpoints, BrainBERT STFT-Large~\citep{wang2023brainbert}, BaRISTA \texttt{parcels\_chans}~\citep{oganesian2025barista}, and MVPFormer-M base~\citep{carzaniga2026mvpformer}.
Omni-iEEG evaluates channel-level predictions or their task-specific aggregates. For the frozen comparison with PHASE-T, we adapt the released foundation models to a single-channel input interface following each model's official protocol, and evaluate their frozen representations with linear probes.
The benchmark provides no electrode coordinates, and its anatomical labels are themselves prediction targets; each model encodes one channel at a time ($C=1$), and BaRISTA receives its released \texttt{UNKNOWN} parcel.
BrainBERT pools the central ten tokens of five-second STFT frames, BaRISTA encodes three-second frames, and MVPFormer filters inputs at 0.5--120\,Hz, resamples them to 512\,Hz, and uses the representation of the last 2,560-sample segment; this band-pass removes most of the 80--500\,Hz HFO band.
All evaluations follow the benchmark's official train/test rows, task-native aggregation, and scoring workflows.
For 60-second rows, BrainBERT and BaRISTA cover the complete
waveform with consecutive frames before averaging their
frame or token representations, and MVPFormer encodes the complete
waveform as twelve consecutive segments, the last of which attends to all twelve.
For HFO events, each event is centered in a single frame, and BaRISTA pools its two central tokens.

Table~\ref{tab:omni-detailed} gives the complete clinical comparison, including
every task-specific specialist published by Omni-iEEG:
LSTM+Attention, adapted from \citet{huang2023emotion};
PatchTST~\citep{nie2023patchtst}; TimesNet~\citep{wu2023timesnet};
eHFO~\citep{monsoor2023hfo}; PyHFO spkHFO and PyHFO-Omni~\citep{zhang2024pyhfo};
CLAP~\citep{laionclap2023}; TimeConv-CNN; SEEG-NET~\citep{wang2022seegnet};
and EEGNet~\citep{lawhern2018eegnet}.

\begin{appendixtable}
\centering
\footnotesize
\setlength{\tabcolsep}{2.5pt}
\renewcommand{\arraystretch}{1.02}
\caption{%
  \textbf{Clinical prediction from events to participant outcomes: full comparison.}
  Rows above the rule are specialist models published by Omni-iEEG; dashes denote tasks that a specialist does not support.
  F1 is macro-F1 and AUC is ROC-AUC. Bold marks the highest performance per column, and underline the highest among frozen encoders.}
\label{tab:omni-detailed}
\resizebox{\textwidth}{!}{%
\begin{tabular}{@{}lrrrrrrr@{}}
\toprule
 & Event & \multicolumn{2}{c}{Window (60 s)} &
 \multicolumn{2}{c}{Anatomy} & \multicolumn{2}{c}{Pathol. brain region} \\
\cmidrule(lr){2-2}\cmidrule(lr){3-4}\cmidrule(lr){5-6}\cmidrule(lr){7-8}
Model & HFO & Ictal & Sleep & Lobe-5 & Region-12 & Channel & Outcome \\
 & F1 & F1 & F1 & F1 & F1 & AUC & AUC \\
\midrule
LSTM+Attention
  & 0.7338 & -- & -- & -- & -- & -- & -- \\
TimesNet
  & 0.7652 & -- & -- & -- & -- & -- & -- \\
PatchTST
  & 0.7726 & -- & -- & -- & -- & -- & -- \\
eHFO
  & -- & -- & -- & -- & -- & 0.6611 & 0.4521 \\
PyHFO spkHFO
  & -- & -- & -- & -- & -- & 0.6557 & 0.4972 \\
PyHFO-Omni
  & 0.8061 & -- & -- & -- & -- & 0.7351 & 0.7438 \\
\addlinespace[2pt]
CLAP
  & -- & 0.9245 & 0.7225 & 0.4750 & 0.3540 & 0.7684 & 0.6770 \\
TimeConv-CNN
  & -- & 0.8533 & 0.7118 & 0.4788 & 0.3087 & 0.8061 & 0.7380 \\
SEEG-NET
  & -- & 0.7526 & 0.6773 & 0.2520 & 0.1081 & 0.7850 & 0.5952 \\
EEGNet (adapted)
  & -- & -- & -- & -- & -- & 0.7468 & 0.5942 \\
\midrule
BrainBERT STFT-Large
  & 0.7937 & 0.8323 & 0.6941 & 0.4054 & 0.2592 & 0.8055 & 0.6017 \\
BaRISTA \texttt{parcels\_chans}, $C{=}1$
  & 0.6050 & 0.7428 & 0.7213 & 0.3006 & 0.1658 & 0.6949 & 0.5739 \\
MVPFormer-M, $C{=}1$
  & 0.4347 & 0.8311 & 0.5404 & 0.2423 & 0.1069 & 0.6604 & 0.5878 \\
\addlinespace[2pt]
PHASE-T, frozen
  & \underline{0.8138} & \underline{0.8958} & \underline{0.7413} & \underline{0.4822} & \underline{0.3398} & \underline{0.8394} & \underline{0.6369} \\
PHASE-T, fine-tuned
  & \textbf{0.8188} & \textbf{0.9327} & \textbf{0.8150} & \textbf{0.5134} & \textbf{0.3572} & \textbf{0.8502} & \textbf{0.7713} \\
\bottomrule
\end{tabular}%
}
\end{appendixtable}

\subsubsection{Direct Physiological-Target Control}
\label{app:phase-t-ablation-full}

Table~\ref{tab:mechanism-ablations-full} compares full PHASE-T with a direct
physiological-target shortcut control that uses no encoder, under the
task-specific fitting and evaluation protocols of the ablations in
Table~\ref{tab:mechanism-ablations}; Full PHASE-T repeats the frozen scores of
Table~\ref{tab:omni}, and Outcome is the participant-level aggregate of the
pathological-channel probe.
The control time-averages the complete 84-dimensional physiological target
for the four 60-s tasks and fits task-matched linear probes. For HFO, whose
570-ms crops are shorter than the support of the low-frequency wavelets, the
control uses the duration-matched 20-dimensional patch-local target; a
60-dimensional high-frequency variant reaches 0.6531.
Full PHASE-T outperforms the encoder-free control on all five tasks.

\begin{appendixtable}
\centering
\scriptsize
\setlength{\tabcolsep}{3pt}
\renewcommand{\arraystretch}{1.05}
\caption{\textbf{Direct physiological-target control across five clinical tasks.}
F1 is macro-F1 and AUC is ROC-AUC. $^\ast$Duration-matched patch-local target.
Bold marks the highest performance per column.}
\label{tab:mechanism-ablations-full}
\begin{tabular*}{\textwidth}{@{\extracolsep{\fill}}lrrrrrrr@{}}
\toprule
 & Event & \multicolumn{2}{c}{Window (60 s)} & \multicolumn{2}{c}{Anatomy} & \multicolumn{2}{c}{Pathol. brain region} \\
\cmidrule(lr){2-2}\cmidrule(lr){3-4}\cmidrule(lr){5-6}\cmidrule(lr){7-8}
Representation & HFO & Ictal & Sleep & Lobe-5 & Region-12 & Channel & Outcome \\
 & F1 & F1 & F1 & F1 & F1 & AUC & AUC \\
\midrule
Full PHASE-T
  & \textbf{0.8138} & \textbf{0.8958} & \textbf{0.7413}
  & \textbf{0.4822} & \textbf{0.3398} & \textbf{0.8394} & \textbf{0.6369} \\
Direct targets (no encoder)
  & 0.6223$^\ast$ & 0.8112 & 0.3089 & 0.2431 & 0.1425 & 0.7293 & 0.5962 \\
\bottomrule
\end{tabular*}
\end{appendixtable}

\subsection{Cross-Institution Transfer}\label{cross-institution-transfer}

For transfer from Omni-iEEG to Hatano~\citep{hatano2026validation}, we evaluate pathological-channel identification using 60-second window representations.
Feature standardization and a logistic probe with $\ell_2$ penalty weight $\lambda=0.01$ are fitted on the Omni-iEEG training windows.
The fitted probe is then applied without adaptation to Tohoku and NCNP.
Target-window probabilities are averaged within recording-channel units, yielding 915 channels from 26 participants at Tohoku and 1,000 channels from 30 participants at NCNP.
Tohoku and NCNP are absent from both pretraining and probe fitting.
Threshold-independent ROC-AUC serves as the evaluation metric.

For comparison, we follow Omni-iEEG~\citep{duan2026omniieeg}
in training TimeConv-CNN, CLAP~\citep{laionclap2023}, and SEEG-NET~\citep{wang2022seegnet} on the same
pathological-channel identification task.
Each model is then applied to the target institutions using
its training-time preprocessing.
The full transfer comparison appears in Table~\ref{tab:external-transfer-detailed}.

\begin{appendixtable}
\centering
\small
\setlength{\tabcolsep}{5pt}
\caption{\textbf{Cross-institution transfer without local labels: full comparison.}
Pathological-channel ROC-AUC at Tohoku and NCNP. Bold marks the highest performance per column.}
\label{tab:external-transfer-detailed}
\begin{tabular*}{\textwidth}{@{\extracolsep{\fill}}llrr@{}}
\toprule
Model & Training regime & Tohoku & NCNP \\
\midrule
TimeConv-CNN & Supervised source model & 0.7434 & 0.4716 \\
CLAP & Supervised source model & 0.7663 & 0.5021 \\
SEEG-NET & Supervised source model & 0.7636 & 0.5528 \\
PHASE-T & Frozen encoder + source probe & \textbf{0.7775} & \textbf{0.6292} \\
\bottomrule
\end{tabular*}
\end{appendixtable}

\subsection{Adaptation from Four Participants}
\label{mayo-fnusa-fine-tuning}

We follow MVPFormer's evaluation protocol~\citep{carzaniga2026mvpformer} on the MAYO and FNUSA release~\citep{nejedly2020multicenter}.
At each institution, the first four participants (release IDs 0--3) are used for training and the rest for testing;
pathological clips are classified against noise and physiological clips, using all clips;
and binary F1 is pooled across test clips.
This yields 21,665 training clips and 133,517 test clips from
20 held-out participants at MAYO, and 33,805 training clips and
159,313 test clips from 10 held-out participants at FNUSA.
The three-second clips are resampled to 1\,kHz.

A linear two-class head is initialized by fitting a logistic probe to
training-participant embeddings with $\ell_2$ penalty weight $\lambda=0.01$. We then train
the head for one epoch and jointly update the encoder and head for five
epochs. AdamW uses encoder and head learning rates of $3\times10^{-5}$
and $10^{-4}$, respectively, weight decay 0.01, an effective batch size
of 32, and 5\% warm-up followed by cosine decay. We use unweighted
cross-entropy, seed 0, and no augmentation. Because released MAYO and
FNUSA clips are standardized to unit variance rather than calibrated
microvolts, inputs are scaled to match the nominal amplitude regime
($s_0=20\,\mu\mathrm{V}$), with checkpoint-owned waveform preprocessing
applied identically during initialization, training, and evaluation.

The final epoch is evaluated with the trained head at probability
threshold 0.5.

We compare against PopT, BaRISTA,
MVPFormer, and NeuroCLUS as reported
in Table~1 of \citet{zheng2026neuroclus}.
MAYO and FNUSA are absent from PHASE pretraining.
Tables~\ref{tab:mayo-fnusa-finetune-full} and
\ref{tab:mayo-fnusa-adaptation-details} give the adaptation
comparators and held-out cohort metrics (ROC-AUC, precision, recall, and
specificity).

\begin{appendixtable}
\centering
\small
\setlength{\tabcolsep}{5pt}
\renewcommand{\arraystretch}{1.08}
\caption{\textbf{Four-participant adaptation: full comparison.}
Binary F1 on held-out participants at MAYO and FNUSA. Bold marks the highest performance per column.}
\label{tab:mayo-fnusa-finetune-full}
\begin{tabular*}{0.70\textwidth}{@{\extracolsep{\fill}}lrr@{}}
\toprule
Model & MAYO & FNUSA \\
\midrule
NeuroCLUS & 0.40\phantom{00} & 0.51\phantom{00} \\
MVPFormer & 0.36\phantom{00} & 0.46\phantom{00} \\
PopT & 0.34\phantom{00} & 0.43\phantom{00} \\
BaRISTA & 0.30\phantom{00} & 0.45\phantom{00} \\
\midrule
PHASE-T & \textbf{0.7337} & \textbf{0.6656} \\
\bottomrule
\end{tabular*}
\end{appendixtable}

\begin{appendixtable}
\centering
\small
\renewcommand{\arraystretch}{1.10}
\caption{\textbf{Four-participant adaptation: held-out cohort metrics.}}
\label{tab:mayo-fnusa-adaptation-details}
\begin{tabular*}{0.70\textwidth}{@{\extracolsep{\fill}}lrr@{}}
\toprule
Quantity & MAYO & FNUSA \\
\midrule
Training participants & 4 & 4 \\
Training clips & 21,665 & 33,805 \\
Test participants & 20 & 10 \\
Test clips & 133,517 & 159,313 \\
\midrule
Binary F1 & 0.7337 & 0.6656 \\
ROC-AUC & 0.9571 & 0.8625 \\
Precision & 0.7023 & 0.5600 \\
Recall & 0.7681 & 0.8201 \\
Specificity & 0.9666 & 0.7780 \\
\bottomrule
\end{tabular*}
\end{appendixtable}

\subsection{Multichannel Auditory Decoding}
\label{braintreebank-auditory-protocol}

We follow BaRISTA's recording split~\citep{oganesian2025barista} and PopT's evaluation protocol~\citep{chau2025popt}: four binary tasks over five-second windows (sentence onset, speech detection, volume, and pitch), within-participant train/validation/test splits, 90 channels per participant and task chosen on validation data, and ROC-AUC averaged over the seven evaluation participants. Signals follow the benchmark's Laplacian referencing scheme and are resampled to 1\,kHz. Both PHASE-ST rows use the spatial-module checkpoint after 5,000 pretraining updates over the frozen PHASE-T teacher at step 60,000.

As in PopT, the frozen row trains only a linear classifier, and the fine-tuned row also updates the spatial module while the temporal encoder stays frozen. The classifier's inverse regularization strength is 0.01. Fine-tuning runs for eight epochs with batch size 32 and AdamW (learning rates $3\times10^{-4}$ for the spatial module and $10^{-5}$ for the classifier, 5\% warm-up, cosine decay), and the state with the highest validation ROC-AUC is evaluated on test. Comparator rows report published results.

\subsection{Seizure Trajectory Extraction}
\label{app:seizure-trajectory-extraction}

For the retrospective seizure analysis, each channel is encoded independently
in 60-s windows with a 10-s stride. Only tokens centered in the middle 10~s
of each window are retained and averaged within each second. Each
time-resolved embedding therefore includes approximately 25--35~s of waveform
context on either side of its timestamp. The trajectories in
Figure~\ref{fig:readings}a--b are aligned to the clinically annotated seizure onset.

\subsection{Cross-Institution Regional Anatomical Correspondence}
\label{app:anatomical-correspondence}

We assess anatomical correspondence across institutions using
16,444 ECoG channels from 135 Detroit participants and 32 UCLA participants
in the eight Region-12 classes with at least 30 ECoG channels at each institution.
For each channel, embeddings are averaged across recording windows,
centered within participants, standardized across features, and
normalized to unit length.

For region-level matching, channels are averaged by anatomical region
within each institution,
and the resulting regional vectors are normalized to unit length.
Each region is compared by cosine similarity with every region
at the other institution.
The same region is the most similar for all 8 UCLA regions
and 6 of 8 Detroit regions,
and it ranks first or second in all 16 comparisons.
Permuting anatomical labels within each institution yields an average
of 1.0 such match per direction;
none of 2,000 permutations reaches the observed 14 ($p=5\times10^{-4}$).
The eight largest similarities between different regions,
from 0.20 to 0.44, all join anatomical neighbors:
motor and somatosensory cortex, superior and inferior parietal cortex,
parietal and occipital cortex, and the two temporal regions.

For channel-level voting (Figure~\ref{fig:readings}c), each of the ten
participants at the other institution with the most similar channels
votes for the region of its most similar channel, weighted by that similarity,
and each region's vote total is divided by the region's frequency
at the institution searched.
Every region's channels are voted into their own region
more often than chance; the balanced accuracy is 0.271
against 0.125 ($p=5\times10^{-4}$).

\section{Spatial Context Analysis and Ablations}
\label{app:st-context}

\subsection{Concurrent-Channel Context Probe}

\begin{figure}[!b]
\centering
\includegraphics[width=0.55\textwidth]{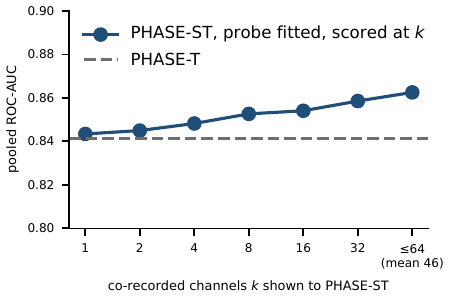}
\caption{\textbf{SOZ identification against panel size.}
Pooled SOZ ROC-AUC as a function of the number of co-recorded channels from
the same participant; the dashed line is PHASE-T.}
\label{fig:st-context-dose}
\end{figure}

Figure~\ref{fig:phasest}a probes whether the frozen spatial stage uses
synchronized channels from the same participant for SOZ
identification. Unlike the single-channel comparison in
Table~\ref{tab:omni}, this analysis constructs channel panels. Linear probes for
the panel-size comparison are fitted and scored on matching panels;
replacement controls also test the fixed same-minute probe. Table~\ref{tab:st-context} lists the conditions, and
Figure~\ref{fig:st-context-dose} shows the pooled ROC-AUC at every panel size.

\paragraph{Panels.}
For this analysis, every training recording (292 recordings, 137 participants) and
every test recording with at least eight labeled channels in a synchronized
window (154 of 237 recordings, 76 participants)
is given two 60-second windows at the labeled rows' own starts, holding
every channel the release marks good; labels are the released labels
(SOZ positive). Training holds 26,286 labeled channel-windows (2,704 SOZ),
test 16,040 (1,446 SOZ). Each channel is encoded once by frozen PHASE-T.
For each panel size $k\in\{1,2,4,8,16,32,64\}$ the channels of a window are
partitioned at random into balanced groups of at most $k$ (at $k=64$ the
mean group holds 46 channels), and the spatial module mixes each group on its
own; three partitions are drawn per window. Substitution arms at the
largest size replace a channel's group-mates by the same channels taken
from the recording's other window (60 seconds apart) or by 63 channels of another test participant's
window; a mixed arm joins each own group of at most 32 channels with 32 channels of another test participant. In the other-minute arm, hardware and reference are unchanged and
the companions remain mutually synchronized, but their activity is no
longer time-aligned with the target channel.

\paragraph{Probes and statistics.}
One logistic probe per size, with $\ell_2$ penalty weight $\lambda=0.01$ on standardized time-averaged
features, is fitted on all three training
partitions and scores the first test partition. The other-minute arm is
scored both by a probe fitted on other-minute training panels and by the
$k=64$ same-minute probe. The mixed and other-participant arms also use the
$k=64$ same-minute probe. Per-participant ROC-AUC is defined for the 52 test
participants with both classes. Paired comparisons use the exact two-sided sign
test on nonzero paired differences and a 10,000-resample participant bootstrap
of the mean and median paired difference.

\paragraph{Results.}
The gain grows with the number of own-participant channels
(Figure~\ref{fig:st-context-dose}). It depends on time-aligned activity: with
matched probes, the same minute exceeds another minute by a mean $+0.033$
[$+0.017$, $+0.051$] per participant (median $+0.006$ [$+0.003$, $+0.030$]; 36 of
50 participants; $p=0.003$), and the other-minute panels score alike under their
own probe and the same-minute probe (mean per-participant difference, own probe
minus same-minute probe, $-0.004$ [$-0.014$, $+0.004$]).
With the same-minute probe held fixed, the participant's own channels exceed 63
channels of another participant by $+0.067$ [$+0.042$, $+0.093$] (median $+0.032$
[$+0.016$, $+0.056$]; 45 of 51; $p=2\times10^{-8}$).

\begin{table}[t]
\centering
\footnotesize
\setlength{\tabcolsep}{4pt}
\caption{\textbf{SOZ identification under different contexts.}
Pooled ROC-AUC and paired per-participant differences ($\Delta$) from the
one-channel condition; the $\le$32-own-plus-32-other-participant row is compared with
the $\le$32-channel same-minute condition. Brackets are 95\% participant-bootstrap
confidence intervals.}
\label{tab:st-context}
\begin{tabular}{@{}llrrr@{}}
\toprule
context given to each channel & probe fitted on & pooled & mean $\Delta$ [95\%] & median $\Delta$ \\
\midrule
PHASE-T, no context & --- & 0.841 & --- & --- \\
1 channel & same & 0.843 & 0 & 0 \\
$\le$8 channels, same minute & same & 0.853 & $+0.018$ [$+0.007$, $+0.032$] & $+0.003$ \\
$\le$32 channels, same minute & same & 0.859 & $+0.041$ [$+0.022$, $+0.061$] & $+0.013$ \\
$\le$64 channels, same minute & same & 0.862 & $+0.048$ [$+0.026$, $+0.071$] & $+0.016$ \\
same channels, another minute & another minute & 0.850 & $+0.015$ [$+0.002$, $+0.029$] & $+0.001$ \\
same channels, another minute & same minute & 0.841 & $+0.019$ [$+0.000$, $+0.040$] & $+0.003$ \\
$\le$32 own $+$ 32 other-participant & same minute & 0.855 & $-0.009$ [$-0.021$, $+0.002$] & $+0.001$ \\
1 own $+$ 63 other-participant & same minute & 0.804 & $-0.019$ [$-0.042$, $+0.004$] & $-0.015$ \\
\bottomrule
\end{tabular}
\end{table}

\subsection{Spatial-Stage Pretraining Ablations}
Both ablated stages in Figure~\ref{fig:phasest}b and c are trained exactly as
PHASE-ST (Appendix~\ref{app:training-inference}), except that one sets the
relation weight to zero and the other restricts each channel's attention to
itself, so that a channel's output depends on its own PHASE-T features alone.
For the stage without cross-channel attention, the gain from adding companions
is therefore zero by construction.
Panel b repeats the panel-size comparison above for each stage. At the largest panels,
the stage without relation targets gains $+0.014$ [$+0.003$, $+0.025$] per
participant; PHASE-ST's context gain is larger by $+0.034$ [$+0.016$, $+0.054$] (95\% intervals
from 10,000-resample paired participant bootstraps).

\subsection{Single-Channel Auditory Probe}
Figure~\ref{fig:phasest}c uses the windows, splits, labels, and 90-channel panel of the
auditory protocol (Appendix~\ref{braintreebank-auditory-protocol}). The spatial stage sees all 90 channels; following BrainBERT~\citep{wang2023brainbert},
a linear probe is fitted to each channel's output.
Each stage's change is taken against the same channel's output from the stage
without cross-channel attention, averaged over the 90 channels and four tasks
within a participant. PHASE-ST exceeds that stage for all seven participants
(mean $+0.054$; smallest difference $+0.038$), and the stage
without relation targets exceeds it by a mean $+0.043$.

\clearpage
\section{Limitations and Future Directions}
\label{app:limitations}

PHASE-T and PHASE-ST are pretrained sequentially,
with the temporal model frozen during spatial pretraining.
Joint single- and multichannel pretraining is a direction
for future study.

Temporal context spans up to one minute.
Longer recordings can be processed in separate windows,
but dependencies across windows are not explicitly modeled.
Extending context toward hour-long recordings will require
efficient sequence modeling and evaluation on tasks that
depend on longer-term changes in neural activity.

Our evaluation covers clinical prediction, cross-institution
transfer, and naturalistic auditory decoding.
Future work will expand pretraining to additional participants
and institutions and evaluate a wider range of recording
contexts and downstream tasks.

\end{document}